\documentclass[journal]{IEEEtran}
\usepackage{xspace,amsmath,amssymb,amsfonts,epsfig,syntonly}
\usepackage{diagbox}
\usepackage{cite,bm,color,url,textcomp}
\usepackage{epstopdf}
\usepackage{empheq}
\usepackage{graphicx}
\usepackage[ruled,linesnumbered]{algorithm2e}
\usepackage{subfigure}
\usepackage[utf8]{inputenc}

\newcommand{\cmmnt}[1]{\ignorespaces}

\long\def\symbolfootnote[#1]#2{\begingroup
\def\thefootnote{\fnsymbol{footnote}}
\footnote[#1]{#2}\endgroup}

\psfull

\allowdisplaybreaks[4]

\begin{document}
\title{
Towards Dynamic Resource Allocation and Client Scheduling in Hierarchical Federated Learning: A Two-Phase Deep Reinforcement Learning Approach
}

\author{Xiaojing Chen, Zhenyuan Li, Wei Ni,~\IEEEmembership{Fellow, IEEE}, Xin Wang,~\IEEEmembership{Fellow, IEEE}, Shunqing Zhang,~\IEEEmembership{Senior Member, IEEE}, Yanzan Sun, Shugong Xu,~\IEEEmembership{Fellow, IEEE}, and Qingqi Pei,~\IEEEmembership{Senior Member, IEEE}
\thanks{Work in the paper was supported by the National Key R\&D Program of China grants 2022YFB2902002 and 2022YFB2902303, the Innovation Program of Shanghai Municipal Science and Technology Commission grant 21ZR1422400, the Program of Hebei Key Laboratory of Advanced Laser
Technology and Equipment grant HBKL-ALTE202401, the National High Quality Program grant TC220H07D, and Shanghai Rising-Star Program. \textit{(Corresponding authors: Shunqing Zhang; Yanzan Sun.)}

X. Chen, Z. Li, S. Zhang, Y. Sun and S. Xu are with the Key Laboratory of Specialty Fiber Optics and Optical Access Networks, Shanghai University, Shanghai 200444, China. Emails:~\{jodiechen, fivears, shunqing, yanzansun, shugong\}@shu.edu.cn.

W. Ni is with the Commonwealth Scientific and Industrial Research Organization (CSIRO), Sydney, NSW 2122, Australia. Email: wei.ni@data61.csiro.au.

X. Wang is with the Key Laboratory of EMW Information (MoE), the Department of Communication Science and Engineering, Fudan University, Shanghai 200433, China. Email: xwang11@fudan.edu.cn.

Q. Pei is with the State Key Laboratory of Integrated Services Networks, School of Telecommunications Engineering, Xidian University, Xi'an 710071, China. Email: qqpei@mail.xidian.edu.cn.}
}

\maketitle

%\markboth{IEEE TRANSACTIONS ON WIRELESS COMMUNICATIONS (TO APPEAR)}%
%{WANG \MakeLowercase{\textit{et al.}}: }

\setcounter{page}{1}
\begin{abstract}Federated learning (FL) is a viable technique to train a shared machine learning model without sharing data. Hierarchical FL (HFL) system has yet to be studied regrading its multiple levels of energy, computation, communication, and client scheduling, especially when it comes to clients relying on energy harvesting to power their operations. This paper presents a new two-phase deep deterministic policy gradient (DDPG) framework, referred to as ``TP-DDPG'', to balance online the learning delay and model accuracy of an FL process in an energy harvesting-powered HFL system. The key idea is that we divide optimization decisions into two groups, and employ DDPG to learn one group in the first phase, while interpreting the other group as part of the environment to provide rewards for training the DDPG in the second phase. Specifically, the DDPG learns the selection of participating clients, and their CPU configurations and the transmission powers. A new straggler-aware client association and bandwidth allocation (SCABA) algorithm efficiently optimizes the other decisions and evaluates the reward for the DDPG. Experiments demonstrate that with substantially reduced number of learnable parameters, the TP-DDPG can quickly converge to effective polices that can shorten the training time of HFL by 39.4\% compared to its benchmarks, when the required test accuracy of HFL is 0.9.
\end{abstract}


\section{Introduction}
% {\color{red}\textbf{ COMMENT:} The purpose of the Introduction section is to provide motivation for the paper. Anything that does not serve this purpose could be removed. For example, the current first paragraph is less relevant. 
% Another key to the Introduction is to quickly tell the reviewers what you do and why. So, the typical structure that I recommend is:

% Paragraph 1) background of the technology and its applications and importance (e.g., your second paragraph), 

% Paragraph 2) key challenges to be addressed in this paper (e.g., resource scheduling in the face of straggler problem), 

% Paragraph 3) limitations of existing literature (high-level summary, since you have separate related work section), 

% Paragraph 4) contribution statement. 
% } 

With the advent of the Internet-of-Things (IoT), pervasively deployed devices are revolutionizing the way we live and work, by producing vast amounts of data \cite{9810265,2022Distributed}. Virtual reality, autonomous driving, and intelligent inference can be enabled by applying deep learning to the rich data. 
As a collaborative decentralized machine learning paradigm, federated learning (FL) is becoming increasingly popular, where IoT devices, namely, clients, download global models from parameter servers, update their local models, and upload the local models for iterative global model updates to the parameter server \cite{mcmahan2017communication}. 
Allowing the model to be updated locally, FL can protect sensitive local data~\cite{2021Distributed} and substantially reduce the burden on network backbones since the models are typically much more lightweight than the data samples~\cite{9060868}. 

The key challenge of FL systems arises from restrictive resources, including finite (and limited) computing power and communication bandwidth, especially in the face of the straggler effect. The straggler effect occurs when each training round only progresses as fast as the slowest client in a synchronous FL process, since the central server has to wait for all clients to complete local training before a global aggregation can take place~\cite{9060868}. The slowest client is known as the straggler, which increases the training latency of FL.
First, due to the limited bandwidth and heterogeneous clients, it is necessary to select appropriate clients to participate in FL and guarantee learning accuracy, while allocating the computation and communication resources efficiently to minimize the learning delay. Densely distributed clients can usually communicate with multiple edge servers, making client association decisions difficult. Moreover, the clients of FL, e.g., sensor nodes in remote areas, are usually powered by finite batteries. Effective energy management is critical to making full use of constrained batteries. As a result, joint optimization of energy, computation and communication resource allocation, and client scheduling is non-trivial to create effective FL. 

Many existing FL studies~\cite{9210812,8761315,9475121,9461628,9261995,konevcny2016federated,8870236,9718315,9509751,8851249,9452072,hamdi2021federated} assumed a cloud server or an edge server as the parameter server, namely cloud- or edge-based FL. However, cloud-based FL can cause a large amount of delay and a high drop-out rate due to communication problems with the cloud server~\cite{9210812}. On the contrary, although edge-based FL can meet benefit from low latency and high reliability, its training performance inevitably declines due to relatively few clients that the edge server can access.
A new client-edge-cloud hierarchical FL (HFL) system was recently proposed in \cite{9148862}, where edge servers serve as intermediaries between the cloud server and clients. Such a hierarchical architecture takes advantage of the cloud- and edge-based FL. On the one hand, the clients can send local models to the nearby edge servers rather than the cloud server for edge aggregation, thereby decreasing the delay and drop rate due to proximate accesses. On the other hand, multiple rounds of edge aggregations are performed before uploading the updated models to the cloud, aggregating a wider coverage of clients and resulting in a more accurate FL model~\cite{Liu2019EdgeAssistedHF}. Yet, making appropriate resource allocation or client scheduling decisions for HFL is challenging due to the complex communication and learning mechanism of HFL~\cite{9880724,luo2020hfel}.

This paper presents a new framework to comprehensively optimize client scheduling (i.e., client selection and association) and resource allocation (including energy, computation, and communication) for HFL. 
We consider the clients equipped with rechargeable batteries and relying on the renewable energy harvested from their environment to power local model training and uploading, which has never been studied in the literature, e.g.,~\cite{luo2020hfel,feng2021min,xu2021dynamic,00925,deng2021share,lim2021decentralized,2021,9488756,abdellatif2022communication,liu2022joint,sun2023joint,wen2022joint,zhang2024device,lim2021dynamic,su2023low,zhao2022drl,zhao2023drl}.
A new Deep Deterministic Policy Gradient (DDPG)-based framework, named two-phase DDPG (TP-DDPG), is developed to
adjust client scheduling and resource allocation adapting to the time-varying wireless channels and renewable energy arrivals, thereby balancing the learning delay and model accuracy of HFL.

Apart from its new consideration of energy harvesting-powered clients and comprehensiveness of the problem tackled, a key contribution of the TP-DDPG framework is that we interpret the client association and bandwidth allocation as part of the environment. An algorithm, named straggler-aware client association and bandwidth allocation (SCABA), is developed to optimize the client association and bandwidth allocation (in the second phase of each iteration of TP-DDPG). SCABA relies on decisions made by a DDPG agent on client selection, transmit power, and CPU configuration (in the first phase), thereby generating rewards for the DDPG agent. 
With the unique and optimal solution of SCABA, the DDPG agent can be trained effectively by interacting with the environment through SCABA. The number of decisions the DDPG agent needs to make is substantially reduced. Consequently, the DDPG can converge rapidly and reliably. 

The key contributions of this paper are summarized as follows. 

\begin{enumerate}
\item[1)] We propose a new energy harvesting-powered HFL system, where clients are equipped with rechargeable batteries of finite capacity and powered by renewable energy sources. The decisions about client scheduling and resource allocation are meticulously optimized, adapting to time-varying channels and renewable energy arrivals. This differs distinctively from the existing works on HFL systems, which have been typically under the prerequisite of persistent power supply for clients. 

\item[2)] We design the new TP-DDPG framework, which comprehensively optimizes client scheduling (i.e., client selection and association) and resource allocation (including CPU frequency, communication bandwidth, and transmit power). %The DDPG outputs decisions on actions, and then evaluates the decisions by evaluating the reward returned by the environment in reaction to the actions. 
% In each iteration/round of our proposed TP-DDPG algorithm, the decisions of the DDPG (in the so-called first phase) are on client selection, transmit power, and CPU configurations. 
Particularly, we divide the optimization decisions into two groups, with a DDPG agent tailored to learn one of the groups, including client selection, transmit power, and CPU configuration.

\item[3)]
We interpret the other group of decisions, including client association and bandwidth allocation, as part of the environment. The new SCABA algorithm is developed to optimize these decisions based on the decisions of the DDPG, thereby generating rewards for the DDPG to refine its policy. 
\end{enumerate}

Extensive experimental results are carried out to assess the proposed TP-DDPG algorithm. 
With its substantially reduced learnable action space, the TP-DDPG can rapidly converge to effective policies that can shorten the training time of HFL by 39.4\% compared to its benchmarks when the required test accuracy of HFL is 0.9 based on the MNIST dataset. TP-DDPG also enables HFL to converge to the highest accuracy of 0.93 on the CIFAR-10 dataset, while the accuracy is lower than 0.9 for all the benchmarks. This two-stage, latency-sensitive FL architecture can be integrated with edge computing infrastructure to perform real-time AI inference and decision-making at the network edge. This is particularly beneficial for applications, where data needs to be processed locally, such as IoT, autonomous vehicles, and augmented reality systems~\cite{xiao2023time}. %the algorithm is able to substantially shorten the learning period while achieving a higher test accuracy for an FL task, compared with the benchmarks. 

Several pioneering works have attempted to optimize resource allocation or client scheduling policies for HFL systems~\cite{luo2020hfel,feng2021min,xu2021dynamic,00925,deng2021share,lim2021decentralized,2021,9488756,abdellatif2022communication,liu2022joint,sun2023joint,wen2022joint,zhang2024device,lim2021dynamic,su2023low,zhao2022drl,zhao2023drl}. These studies usually considered a static cloud aggregation process (or global iteration), for which fixed resource allocation and/or client scheduling decisions were made off-line, e.g., using game theory~\cite{lim2021decentralized,lim2021dynamic}, convex approximation~\cite{luo2020hfel, liu2022joint, wen2022joint}, and alternating optimization~\cite{sun2023joint}. These studies, in general, cannot adapt to dynamically changing HFL systems with time-varying wireless channels. While deep reinforcement learning (DRL) has the potential to adapt to environmental changes, the existing DRL-based solutions, i.e.,~\cite{zhao2022drl,zhao2023drl}, were designed to perform client scheduling and resource allocation one-off (or off-line) and would suffer from limited scalability, if employed online, due to a direct collection of all actions in their action spaces and consequently relatively large action spaces. 
% Finer-grained decision-making would be critical to dealing with the dynamically changing nature of practical HFL systems. 
On the other hand, energy harvesting technology is a key enabler of energy-efficient and self-sustainable HFL systems where the clients are powered by constrained batteries, but it has yet to be considered in these HFL systems. The incorporation of energy harvesting-powered clients would also require client scheduling and resource allocation to adapt to changing energy arrivals, in addition to the time-varying wireless channels.

Unlike the existing studies, in this paper, we consider a new HFL system with clients equipped with rechargeable batteries of finite capacity and relying on the renewable energy harvested from their environment to power local model training and uploading. The proposed TP-DDPG algorithm optimizes client scheduling and resource allocation (including bandwidth allocation, and the CPU frequencies and transmit powers of the clients) online during each round of edge aggregation, adapting to the time-varying system environment.

We note that the convergence speed and stability of a reinforcement learning (RL) algorithm depend heavily on the sizes of its action spaces. The larger the action space is, the more difficulty the RL algorithm can have in converging. As a matter of fact, the convergence time could increase exponentially as the action space grows, if it indeed converges. In many other cases, such an RL algorithm may not converge or even diverge. 
% This phenomenon is known as ``the curse of dimensionality'' in the context of RL.
In light of this, the proposed TP-DDPG algorithm judiciously segments the action space of the problem at hand and interprets one part of the action space as part of the environment for the other part of the action space. As a consequence, the remaining action space that requires the attention of DDPG is substantially reduced, leading to enhanced convergence and reliability. 
This can be critical to the efficiency and scalability of HFL systems, when the network is large with more clients involved.

The rest of this paper is organized as follows. An overview of the related works is given in Section II. A model of the HFL system is presented in Section III. A formulation of the problem is provided in Section IV. We elaborate on the new TP-DDPG framework in Section V and evaluate it experimentally in Section VI. The paper concludes with Section VII.

%在三层联邦学习下进行用户选择和资源分配
%将联邦学习与EH相结合
%考虑了信道增益和能量到达率的动态变化。

\label{sec:intro}

\section{Related Works}

%{\color{red}\textbf{ COMMENT:} The purpose of the related work section is to show the gap in the literature in regards to the problem solved in this paper. So, please group the papers with similar strengths and similar weaknesses in a paragraph, and clarify what the weakness is in a polite way as I suggested below.} 

Since it was proposed in \cite{mcmahan2017communication}, FL has received widespread attention for its capability of collaboratively training a shared ML model in a privacy-preserving manner. FL still faces challenges arising from practical implementation, including energy efficiency, communication efficiency, model accuracy, and the straggler effect \cite{9060868, 9220780}. A number of existing works endeavored to address these challenges from different perspectives, 
%{\color{red}{\textbf{ COMMENT:} Please also mention the common limitation of the following papers (in one sentence) in regards to the problem that you try to solve here}}. 
typically under the settings of edge- or cloud-based FL, e.g.,~\cite{9475121,9461628,9261995,konevcny2016federated,8870236,9718315,8761315,9509751,8851249,9210812,9452072}. 
The studies in \cite{9475121,9461628,9261995} improved the energy efficiency of FL
without leveraging battery (dis)charging. By optimizing simultaneous communication and computation resource allocation in NOMA and TDMA, Mo and Xu~\cite{9475121} minimized the total energy consumption at edge clients. For FL networks with CPU-GPU platforms, an energy-efficient resource allocation algorithm was proposed in \cite{9461628}, allowing users to assign resources according to their individual needs. Considering heterogeneous computing and power resources, optimal resource allocation was investigated in \cite{9261995}. %In \cite{8664630}, a control algorithm was proposed to determine the best trade-off between local update and global parameter aggregation to minimize the learning loss under resource budgets, including delay, energy and communication bandwidth. 

Studies were conducted to improve the efficiency of FL's communications~\cite{konevcny2016federated,8870236,9718315}. 
%{\color{red}However, .....\textbf{ COMMENT:} Please also mention the common limitation of the following papers (in one sentence) in regards to the problem that you try to solve here.}
These works focused on model compression or communications, and resource allocation or client scheduling was not considered. Compressing model information was introduced and exhibited significant improvement in communication efficiency with minimum impact on training accuracy in \cite{konevcny2016federated}. To accelerate global model aggregation, a broadband analog aggregation (BAA) scheme was designed to take advantage of multi-access channels' waveform superposition property in \cite{8870236}. Along this vein, a unit-modulus analog receive beamforming design for multi-antenna systems was proposed in \cite{9718315}.

% A digital version of the BAA, named one-bit broadband digital aggregation (OBDA), was later developed in \cite{9718315} to deal with the difficulty of deploying the BAA in modern wireless systems.

To improve model accuracy and alleviate the straggler effect, client scheduling strategies were investigated in energy- and/or latency-aware FL systems \cite{8761315,9509751,8851249,9210812,9452072}.  
%{\color{red}However, .....\textbf{ COMMENT:} Please also mention the common limitation of the following papers (in one sentence) in regards to the problem that you try to solve here.}  
None of these works integrated energy harvesting techniques for computation-intensive clients or considered a holistic optimization of client scheduling, bandwidth allocation, and the CPU frequencies and emission power of the clients. A heuristic greedy algorithm was suggested in \cite{8761315} to choose the largest number of clients within a predefined timeframe for training. By latency-aware client selection and resource allocation, Yu \textit{et al.} \cite{9509751} minimized energy consumption while maximizing selected clients. The impact of different scheduling policies on the performance of FL was studied in \cite{8851249}. Client scheduling strategies were also investigated  considering packet errors \cite{9210812} and imperfect channel state information \cite{9452072}, respectively.

% Regarding the number of training rounds and participant clients per round, an upper bound was derived for the training loss of FL in \cite{8851249}. With the aim of maximizing model accuracy within a given learning delay, the bound was used to select clients and allocate resources. 

%All the above works assumed edge-based or cloud-based FL. Yet, HFL has been considered a superior FL framework with a lower delay and better convergence. 
%{\color{red}However, .....\textbf{ COMMENT:} Please also mention the common limitation of the following papers (in one sentence) in regards to the problem that you try to solve here.} 

Recently, several works have attempted to optimize resource allocation and/or client scheduling in HFL systems \cite{luo2020hfel,feng2021min,xu2021dynamic,00925,deng2021share,lim2021decentralized,2021,9488756,abdellatif2022communication,liu2022joint,sun2023joint,wen2022joint,zhang2024device,lim2021dynamic,su2023low,zhao2022drl,zhao2023drl}. However, the strategies developed in the works are static within a cloud aggregation process, failing to adapt to a dynamically changing environment between different edge aggregations \cite{luo2020hfel,feng2021min,xu2021dynamic,00925,deng2021share,lim2021decentralized,2021,9488756,abdellatif2022communication,liu2022joint,sun2023joint,wen2022joint,zhang2024device,zhao2022drl}.
Luo \textit{et al.} \cite{luo2020hfel} and Zhang \textit{et al.} \cite{zhang2024device} collectively optimized the CPU frequency of clients, bandwidth allocation, and client association to effectively reduce the energy consumption and latency of HFL. Feng \textit{et al.} \cite{feng2021min} embarked on the cost minimization of individual clients, instead of all clients, and formulated a min-max problem to minimize the worst-case cost of a client. To improve the training performance, Xu \textit{et al.} \cite{xu2021dynamic} proposed to select clients with more important local updates, and Qu \cite{00925} maximized the number of successful participating clients without dropping out.
Deng \textit{et al.} \cite{deng2021share} minimized the communication cost by choosing a subset of distributed nodes as the edge aggregator and making decisions on distributed node association. Considering running multiple FL tasks simultaneously, Lim \textit{et al.} \cite{lim2021decentralized} developed a framework for resource allocation and incentive mechanism design based on evolutionary game theory. Optimal client associations were also studied to minimize the number of edge-cloud communication rounds \cite{2021}, loss function \cite{9488756}, Kullback–Leibler divergence (KLD) of data distributions~\cite{abdellatif2022communication}, or learning latency~\cite{liu2022joint}.
A two-layer algorithm based on genetic algorithm and alternating optimization was proposed to minimize the
weighted sum of the optimality gap and overall latency in \cite{sun2023joint}. A
joint helper scheduling and wireless resource allocation scheme was proposed in \cite{wen2022joint} to capture the importance of weighted gradient. Dynamic resource allocation was studied in \cite{lim2021dynamic,su2023low,zhao2023drl}, which however overlooked the energy consumption or latency of training.

%The resource allocation and/or client scheduling strategies developed for HFL in \cite{luo2020hfel,xu2021dynamic,feng2021min,00925,deng2021share,lim2021decentralized,2021,9488756} are static within a cloud aggregation process, failing to adapt to dynamically changing environment between different edge aggregations. 

\begin{table}[t]\addtolength{\tabcolsep}{-2pt}\renewcommand\arraystretch{1.2}
 \centering
 \caption{Notation and Definition}\label{tab:no}
\begin{tabular}{p{2cm}|p{6cm}}
 \hline
 \textbf{Notation }& \textbf{Definition} \\ \hline
 ${\mathcal{K}}$ & Set of edge servers \\
  ${{\mathcal N}}$ & Set of clients\\
${\mathcal{D}}_{n}$ & Client $n$’s training data set \\ $\alpha _n^t$   & Selection decision of client $n$ \\ 
$U$ &  Utility function \\ $p_n^{t, com}$ & Transmission power of client $n$  \\ 
$\overline{\boldsymbol{{\omega}}}_k^{t-1}$ & Edge model aggregated by edge server $k$ at the $t$-th edge aggregation \\  $\boldsymbol{\omega}_n^i$ & Local model of client $n$ at the $i$-th local iteration \\ 
$\overline{\boldsymbol{\omega}}^{m}$ &  Global model aggregated by the cloud at the $m$-th cloud aggregation \\ $R$ & Number of cloud aggregation rounds of an FL task \\ 
${R_1}$ & Number of edge aggregation rounds in one cloud aggregation  \\ ${R_2}$ & Number of local iterations in one edge aggregation \\
${\Omega}^{t}$ & Set of clients selected in the $t$-th edge aggregation round  \\  ${{\mathcal{Z}}_k^t}$ & Set of clients associated with edge server $k$  \\ 
$f_n^t$ & CPU frequency of client $n$ for local training \\ $b_{nk}^{t}$ & Ratio of bandwidth allocated to client $n$ from edge server $k$ \\ 
$T_n^{t, com},E_n^{t, com}$ & Communication delay and energy respectively for
client $n$ to upload local model \\ $T_n^{t, cmp},E_n^{t, cmp}$ & Computation delay and energy respectively
of $R_2$ local iterations of client $n$ \\  
$T_g$ & Total latency of edge model upload and global model aggregation and sharing \\ $T_k^{t}$ & Total latency of edge server $k$ in the $t$-th edge aggregation
round \\  
$T^{t}$ &   Latency of completing the $t$-th edge aggregation across all edge servers \\  $E_{n}^{t}$ & Battery energy level of client $n$ at the beginning of the $t$-th edge aggregation round  \\ 
$E_{n, c}^{t}$ & Battery energy level of client $n$ at the end of ``on" time \\  ${T_c^m}$ & Delay of the $m$-th cloud aggregation  \\ 
$E_{n, ho}^{t},E_{n, ho}^{t}$ & Harvested energy during  ``on'' time and ``idle" time respectively \\ ${U_n}$  &  Importance of the local model of client $n$  \\ 
${\tau_{n,t}}$ & Latest edge aggregation round when client $n$ was selected
before the current $t$-th round \\ $F$ & Maximum
number of edge aggregation rounds between two consecutive selections of a client \\ 

 \hline
 \end{tabular}
 \end{table}

\label{sec:rela}

\section{System Model}
\subsection{HFL Framework}

\begin{figure}[t]
\centering
\includegraphics[scale=0.4]{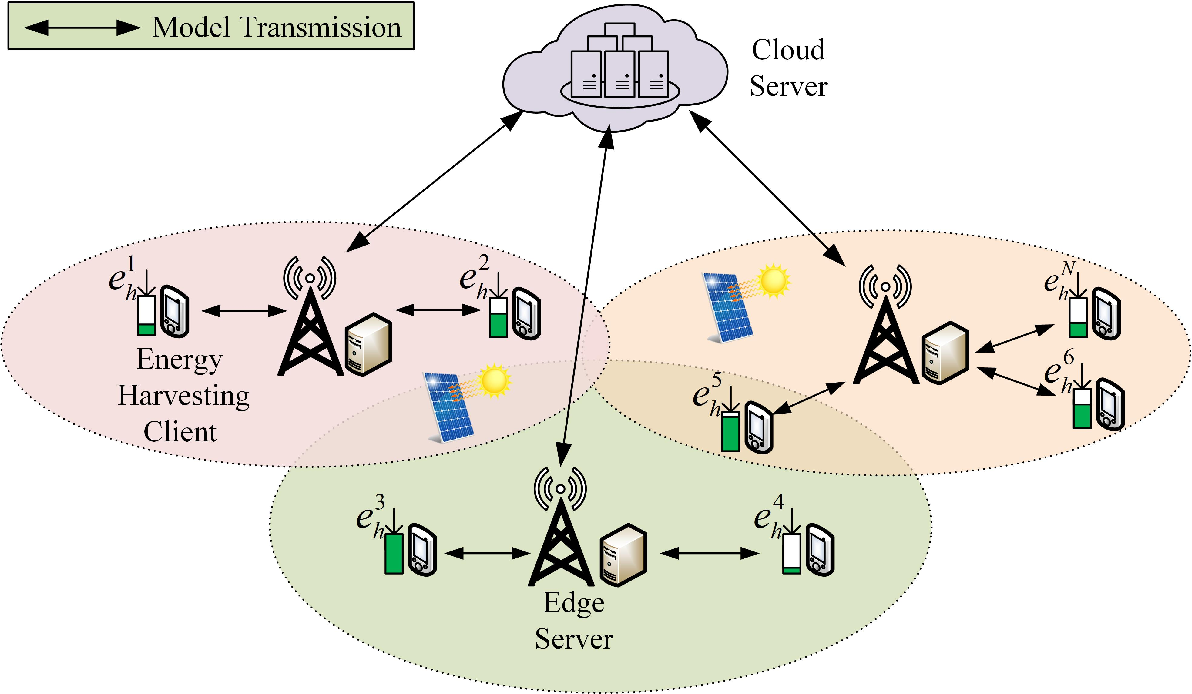}
\caption{Illustration of the HFL system. }
\label{fig:model}
\end{figure}

The considered HFL system comprises a cloud server, a set of edge servers $ {\mathcal{K}}:=\{1, \ldots, K\}$ co-located with $K$ base stations (BSs), and a set of clients ${{\mathcal N}}:=\{1, \ldots, N\}$ powered by renewable energy sources, as illustrated in Fig. \ref{fig:model}. $N\geq K$. Each client $n\in{\mathcal N}$ has a local dataset ${\mathcal{D}}_{n}$ with $|\mathcal{D}_n|$ data samples, where $\mathbb{|\cdot|}$ stands for cardinality. For each dataset ${\mathcal{D}}_{n} = \{ {{x}_{n,d}},{y_{n,d}}\} _{d = 1}^{|\mathcal{D}_n|}$, ${x}_{n,d}$ is the $d$-th input data at client $n$, and ${y_{n,d}}$ is the corresponding label. Let $\boldsymbol{\omega}$ be the parameters of the global FL model, and $f(\boldsymbol{\omega},{{x}_{n,d}},{y_{n,d}})$ be the loss function of the FL with the input ${x}_{n,d}$ and the labeled output ${y_{n,d}}$. The local loss function of client $n$ is ${F_n}(\boldsymbol{\omega} ) = \frac{1}{{{|\mathcal{D}_n|}}}\sum\limits_{d = 1}^{{|\mathcal{D}_n|}} {f(\boldsymbol{\omega} ,{{x}_{n,d}},{y_{n,d}})}$. The objective of training the global FL model is to minimize the following global loss function, i.e.,
\begin{equation}
\mathop  F({\bf{\boldsymbol{\omega} }}) = \frac{1}{\mathcal{|D|}}\sum\limits_{n = 1}^N {{|\mathcal{D}_n|}{F_n}(\boldsymbol{\omega} )}, 
\label{minF}
\end{equation}
where $\mathcal{\mathcal{D}} = \bigcup\limits_{n \in {{\mathcal N}}}{\mathcal{D}}_n$ collects all data samples of all clients. 

To minimize $F({\bf{\boldsymbol{\omega}}})$ in (\ref{minF}) without sharing datasets among the clients and servers, the HFL system adopts an iterative learning protocol. We assume that a total of $R$ rounds of cloud aggregations (i.e., at the cloud server) are required to complete an FL task. Suppose that $\overline{\boldsymbol{\omega}}^0$ is a randomly initialized global model. At the beginning of the $m$-th cloud aggregation round, $m\in[1,R]$, the cloud server sends the global model $\overline{\boldsymbol{\omega}}^{m-1}$ to all edge servers. Then, the HFL system proceeds with $R_1$ rounds of edge aggregations (i.e., at the edge servers) before a cloud aggregation is carried out at the server.

In the $t$-th edge aggregation round that is between the $(m-1)$-th and the $m$-th cloud aggregations, i.e., $t \in [(m - 1){R_1} + 1, m{R_1}]$, the cloud server first requests system information and makes resource allocation and client scheduling decisions. Based on these decisions, each edge server $k \in {\mathcal{K}}$ broadcasts its model $\overline{\boldsymbol{{\omega}}}_k^{t-1}$ through its associated base station to its clients $n \in {{{\mathcal{Z}}_k^t}}$, where ${{\mathcal{Z}}_k^t}$ is the set of clients associated with edge server $k$ in the $t$-th edge aggregation round. Clearly, $\overline{\boldsymbol{\omega}}_k^{t - 1} = \overline{{\boldsymbol {\omega}}}^{m - 1}$ if $t = (m - 1){R_1} + 1$. At any moment, an active client can associate with only an edge server, i.e., 
\begin{align}
& {\mathcal{Z}}_i^{t} \cap {\mathcal{Z}}_k^{t} = \emptyset,~~i \ne k~\text{and}~i,k \in {{\mathcal K}}. \label{asso2}
\end{align} 
The set of clients selected in the $t$-th edge aggregation round is ${\Omega}^{t} =\cup_{k \in {{\mathcal K}}}{\mathcal{Z}}_k^{t}$. Let a binary decision variable $\alpha _n^t$ indicate whether client $n$ is selected to be active in the $t$-th edge aggregation round. $\alpha _n^t=1$, if $n \in {{\Omega}^{t}}$; $\alpha _n^t=0$,  otherwise.

Upon receiving an edge model, client $n$ starts to update its local model for $R_2$ rounds. At the $i$-th local iteration that is within the $t$-th edge aggregation round, i.e., $i\in [(t - 1){R_2} + 1,t{R_2}]$, client $n$ performs its local update:
\begin{equation}
   \boldsymbol{\omega}_n^i = \boldsymbol{\omega}_n^{i - 1} - \eta \nabla {F_n}(\boldsymbol{\omega}_n^{i - 1}),
\end{equation}
where $\eta$ is the learning rate and $\nabla {F_n}(\boldsymbol{\omega}_n^{i - 1})$ is the gradient of ${F_n}(\boldsymbol{\omega}_n^{i - 1})$. Clearly, $\boldsymbol{\omega}_n^{i - 1} = \overline{\boldsymbol{\omega}}_k^{t - 1}$ if  $i = (t - 1){R_2} + 1$. 

After the local update, an orthogonal frequency division multiple access (OFDMA) protocol is employed for the selected clients to upload their local models with a typical size of $10^4 \sim 10^{12}$ bits\cite{villalobos2022machine}.
Each edge server provides a bandwidth of $B$ to its selected clients. 
Edge server $k$ receives and averages the updated model parameters $\{\boldsymbol{\omega}_n^{t{R_2}}, \forall n\in {{\mathcal{Z}}_k^t}\}$ from its associated clients.
% \begin{equation}
% {\boldsymbol{\omega} _k^t} = \frac{{\sum_{n \in {S_k^t}} {|{D_n}|{\boldsymbol{\omega} _n^{R_2}}} }}{{\sum_{n \in {S_k^t}} {|{D_n}|} }} .
% \end{equation}

Consider an importance-oriented weighting proposed in~\cite{lai2021oort} for model aggregation, e.g., to improve model convergence performance when the clients have heterogeneous data distributions.
The importance weights can be calculated at the clients based on locally available information, i.e., the edge model of the $(t-1)$-th round ${\overline{\boldsymbol{\omega}}_k^{t-1}}$, and the local dataset ${\mathcal{D}}_{n} = \{ {{x}_{n,d}},{y_{n,d}}\} _{d = 1}^{|\mathcal{D}_n|}$, as given by [31, eq.(1)]
\begin{equation}
{U_n} = |{\mathcal{D}_n}|\sqrt {\frac{1}{{|{\mathcal{D}_n}|}}{{\sum\limits_{d \in {{\mathcal{D}}_n}} {f({\overline{\boldsymbol{\omega}}_k^{t-1}},{{x}_{n,d}},{y_{n,d}})} }^2}}. 
\end{equation}

The aggregated edge model is given by
%Considering the clients may have heterogeneous data distributions, we apply an importance-oriented weight for the model parameter of client $n$, i.e., ${\boldsymbol{\omega}_n^{tR_2}}$, for edge aggregation~\cite{lai2021oort}: 
\begin{equation}\label{imp-weight}
{\overline{\boldsymbol{\omega}} _k^{t}} = \frac{{\sum_{n \in {{\mathcal{Z}}_k^t}} {{U_n}{\boldsymbol{\omega} _n^{tR_2}}} }}{{\sum_{n \in {{\mathcal{Z}}_k^t}} {{U_n}} }}.
\end{equation}
% where  
% \begin{equation}
% {U_n} = |{\mathcal{D}_n}|\sqrt {\frac{1}{{|{\mathcal{D}_n}|}}{{\sum\limits_{d \in {{\mathcal{D}}_n}} {f({\overline{\boldsymbol{\omega}}_k^{t-1}},{{x}_{n,d}},{y_{n,d}})} }^2}}
% \end{equation}
% denotes the importance of the local model of client $n$ \cite{lai2021oort}. The importance weights are assumed to be known {\it a-priori} since the importance weights of the local models in the $t$-th edge aggregation round can be calculated, e.g., at an edge server, based on the edge model of the $(t-1)$-th round. 
It is worth mentioning that the algorithm proposed in this paper is not limited to a particular scheme, and can apply to other aggregation methods, e.g., the sample number-based weighting~\cite{luo2020hfel}\footnote{Poisoning attacks on FL (including model poisoning and data poisoning) and defense mechanisms, such as model analysis, Byzantine robust aggregation, and verification-based methods, have been studied~\cite{xia2023poisoning}, which are beyond the scope of this paper.}.

When $t=mR_1$, all edge servers deliver their updated edge models $\{\overline{{\boldsymbol{\omega}}} _k^{mR_1}, \forall{k}\}$ to the cloud server for the $m$-th global model aggregation, i.e.,
\begin{equation}
\overline{\boldsymbol{\omega}}^m  = \frac{{\sum_{k \in  \mathcal{K}} {{{D}_k^m}{\overline{\boldsymbol{\omega}}^{mR_1} _k}} }}{{\sum_{k \in { {\mathcal K}}} {{{D}_k^m}} }}, 
\end{equation}
where ${{D}_k^m}={\sum\limits_{t=(m-1)R_1+1}^{mR_1}  {{|\mathcal{D}_k^t|}}}$ and $\mathcal{D}_k^t=\bigcup\limits_{n \in {{\mathcal{Z}}_k^t}} {{\mathcal{D}_n}}$. This concludes a cloud aggregation round and repeats until the global model parameter $\overline{{\boldsymbol{\omega}}}^m$ meets an accuracy requirement.

The proposed HFL framework works in a synchronous fashion, as it requires all edge servers to commence edge model aggregation only after receiving the local models from all its associated clients, and the cloud server to commence global model aggregation only after receiving the edge models from all edge servers. Synchronous FL has the advantage of ensuring that the global model is updated with the most recent information from the clients before aggregation, generally leading to fewer aggregation rounds and better convergence of the global training~\cite{zhang2023fedmds}. This helps avoid the stale model problem where slow clients may use outdated local models. Synchronous FL is suitable for network infrastructures synchronized, maintained, and operated closely by a network operator, such as the edge networks considered in this paper. The widely adopted network time protocol (NTP) can provide the required synchronization~\cite{NTP}. Specifically, an epoch of local training could last for seconds or even minutes at individual clients. By contrast, the network synchronization protocols, such as NTP, synchronize network devices typically within milliseconds~\cite{NTP}. The required synchronization of model training and aggregation can be achieved by evaluating the processing speed of individual clients or edge servers, and selecting the slowest to specify the durations of aggregation rounds.

On the other hand, distinctively different from synchronous FL \cite{luo2020hfel} and FL with so-called flexible aggregation \cite{pan2022adaptive}, the aggregations in asynchronous FL start once an edge (or cloud) server receives a prespecified number of local (or edge) models, instead of all local (or edge) models. The formulation and resource allocation of asynchronous FL are substantially different from synchronous FL and beyond the scope of this paper.

\subsection{Computation Model}

Local updates can be performed using stochastic gradient descent (SGD). A client randomly selects $M$ samples from the local training data for its local update. Let $c_n$ denote the required CPU cycles to process a bit of data at client $n$. Suppose that all samples are of the same size, $\beta$ (in bits). During a local iteration, the CPU consumes ${c_n}M\beta$ CPU Cycles at client $n$.

Let $f_n^{t} \in [0,f_n^{\max}]$ denote the CPU frequency  assigned to the local model training at client $n$ before the $t$-th edge aggregation. The total CPU frequency assigned to client $n$ should not be larger than $f_n^{\max}$. The computation latency for client $n$ to perform $R_2$ rounds of local model updates can be expressed as:
\begin{equation}
T_n^{t, cmp} = \alpha _n^t\frac{{{R_2}{c_n}M\beta}}{{f_n^{t}}}.    
\end{equation}
The energy consumption of the local model updates during $T_n^{t, cmp}$ is \cite{9461628}:
\begin{equation}
E_n^{t, cmp} = p_n^{t, cmp}T_n^{t, cmp}, 
\end{equation}
where $p_n^{t, cmp} = {u_n}{(f_n^t)^3}$ is the power consumption of local computing at client $n$, and ${u_n}$ is a constant depending on the effective capacitance coefficient of its computing chipset. 

\subsection{Communication Model}
Suppose that client $n$ is associated with edge server $k$ in the $t$-th edge aggregation round. Let $b_{nk}^{t}\in [0, 1]$ denote the proportion of the edge server's bandwidth allocated to client $n$ at the round, and $p_n^{t, com}\in[0,p_n^{\max}]$ denote the transmission power of client $n$. According to Shannon's theorem, the achievable transmit rate of client $n$ is given by 
\begin{equation}
   r_{nk}^{t} = b_{nk}^{t}B{\log _2}(1 + \frac{{p_n^{t, com}h_{nk}^{t}}}{\psi^t }),
\end{equation}
where $\psi^t$ is the receiver noise power of the BS, and ${h_{nk}^{t}}\in \mathcal{R}^+$ is the channel gain from client $n$ to edge server $k$. $\mathcal{R}^+$ denotes the set of positive real values. Since different edge servers (more explicitly, their associated BSs) operate at different, non-overlapping channels and the OFDMA protocol is adopted by the selected clients in each of the channels, there is no interference between the edge servers and between the clients in the uplink.  
%{\color{red}\textbf{COMMENT:} We may want to clarify the edge servers use different frequency bands to communicate with the clients, so that there is no interference between the edge servers or between the clients in the uplink. }

Let $\zeta$ denote the size of the local models produced by the clients. The communication delay for uploading ${\boldsymbol{\omega}_{n}^{tR_2}}$ from client $n$ to edge server $k$ is given by
\begin{equation}
 T_n^{t, com} = \alpha _n^t\frac{\zeta }{{r_{nk}^{t}}}.   
\end{equation} 
%Given the transmission time and power, 
The energy cost for uploading ${\boldsymbol{\omega}_{n}^{tR_2}}$ from client $n$ is:
\begin{equation}
 E_n^{t, com} = p_n^{t, com}T_n^{t, com}.
\end{equation}

Under a synchronous FL framework, the total latency of edge server $k$ in the $t$-th edge aggregation round depends on the slowest client under the edge server and is given by
\begin{equation}
T_k^{t} = \mathop {\max}\limits_{n \in {\mathcal{Z}}_k^{t}} (T_n^{t, cmp} + T_n^{t, com}+T_e),   
\end{equation}
where $T_e$ is a constant accounting for the downlink model sharing  and the model aggregation of the edge servers \cite{luo2020hfel}.

The $(t+1)$-th edge aggregation round does not start until all edge servers complete their $t$-th edge aggregations. This is because the client association and resource allocation may change between edge aggregation rounds. The latency of the $t$-th edge aggregation round is thus given by: 
\begin{equation}
{T^{t}} = \mathop {\max }\limits_{k \in {\mathcal K}} T_k^{t}. 
\end{equation}

When $t=mR_1$, the cloud collects the updated models from all edge servers and combines them into a global model.
The delay of the
$m$-th cloud aggregation is
\begin{equation}
    {T_c^m} = \sum\limits_{t = (m - 1){R_1}+1}^{m{R_1}} {{T^t}}+T_g,
\end{equation}
Here, $T_g=T_u+T_c$. In particular, $T_u$ accounts for the constant latency for the edge servers to transmit their aggregated models to the cloud server; $T_c$ contains the constant model aggregation delay and global model sharing delay of the cloud server \cite{abdellatif2022communication}. 

%{\color{red}\textbf{COMMENT:} I don't see why (13) is needed since each edge server just does its own job and why it needs to synchronize with other servers. You may argue that the user association and resource allocation may change between edge aggregation rounds, e.g., a client may switch between one edge server to another. If this is the case, please clarify it after the red text above (13).}

\subsection {Energy Harvesting and Battery Models}

The clients each have a rechargeable battery with a capacity of ${E^{\max}}$, and are powered by renewable energy. %The energy harvesting is modeled as a Poisson process with a different mean $e_n^h\in [e_h^{\min },e_h^{\max}]$ at different clients. 
Let $E_n^{t}$ indicate the battery charging level of client $n$ when the $t$-th edge aggregation round starts; $E_n^{1}$ is the initial battery energy level of client $n$. %which is uniformly distributed in $\left[ 0, {E^{\max }} \right]$. 

%{\color{red}\textbf{COMMENT:} Don't keep adding new notations like $T_{start}^t$, which makes the paper really hard to follow. Also, the meaning of $t$ is rather confusing here, an index to time or round? 
%I suggest using a unified timeline to describe, as suggested earlier on the previous page.}

\begin{figure}[h]
\centering
    \includegraphics[scale=0.67]{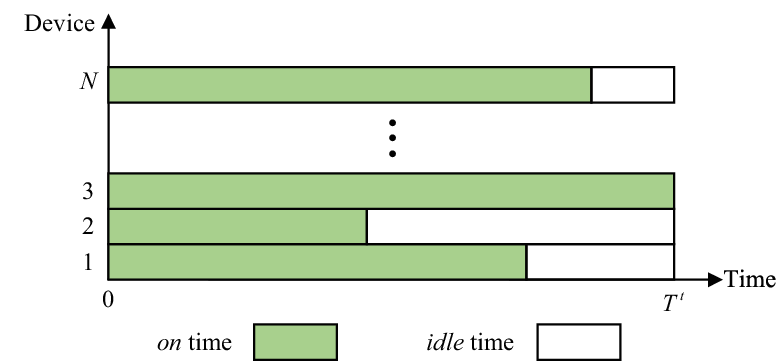}
\caption{The time interval of the $t$-th edge aggregation.}
\label{fig:time_interval}
\end{figure}

We divide the $t$-th edge aggregation round into: i) an \textit{on} time from 0 to $T_n^{t, cmp} + T_n^{t, com}$, and ii) an  \textit{idle} time from $T_n^{t, cmp} + T_n^{t, com}$ to $T^{t}$; see Fig. \ref{fig:time_interval}. A client keeps harvesting energy during both the ``on'' and ``idle'' times. The energy harvesting process follows a Poisson distribution with different means $e_n^h\in [e_h^{\min },e_h^{\max}]$ at different clients.
%If client $n$ is selected in the $t$-th edge aggregation round, it harvests energy during the entire $[0,T^{t}]$. 
The amounts of energy harvested during the ``on'' time and the ``idle'' time are $E_{n, ho}^{t}$ and $E_{n, hi}^{t}$, respectively. 

Then, the battery energy level of client $n$ at time $T_n^{t, cmp} + T_n^{t, com}$ is
\begin{equation}
E_{n, c}^{t} =\min \{  E_n^{t} + E_{n, ho}^{t} - E_n^{t, cmp} - E_n^{t, com}, E^{\max} \},   
\end{equation}
where $E_n^{t} + E_{n, ho}^{t} \geq E_n^{t, cmp}+ E_n^{t, com}$ ensures that the energy initially stored in the battery and collected during the ``on'' time should be larger than the energy consumption over the ``on'' time.
When the $(t+1)$-th edge aggregation round starts, the battery level amounts to:
\begin{equation}
 E_n^{t+1} = \min \{ E_{n, c}^{t} + E_{n, hi}^{t}+E_{g}^{t}, {E^{\max}}\},
\end{equation}
where $E_{g}^{t}$ denotes the harvested energy from time $T^t$ to $(T^t+T_g)$, if a cloud model aggregation occurs after the $t$-th edge aggregation round; or $E_{g}^{t}=0$, otherwise.

% Considering the energy stored in the battery and  collected during work time should cover the energy consumption  over the work time, the battery energy level at time $T_n^{t, cmp} + T_n^{t, com}$  can be defined as:
% \begin{equation}
%  E_{n, c}^{t} =\min \{  [ E_n^{t} + E_{n, ho}^{t} - E_n^{t, cmp} - E_n^{t, com}]^{+},E^{max} \} .    
% \end{equation}
% where $[X]^{+}=\max \{X,0 \}$. Then, the battery energy $E_n^{t+1}$ can be updated as:

\label{sec:model}

\section{Problem Formulation}
To accomplish an FL task with a reasonable learning delay and model accuracy, we optimize resource allocation and client schedule in this section.
Given the total number of local iterations of an FL task (i.e., $I:=RR_1R_2$) and fixed dataset size, $|\mathcal{D}_n|, \forall n$, we can show that the expected global gradient deviation is upper bounded by 
{\small \begin{equation}
\begin{aligned}
   &\frac{1}{I}\sum_{i=1}^{I} \mathbb{E}\Vert\nabla{F}(\boldsymbol{\overline{\boldsymbol{\omega}}^{m_i}})\Vert^2\leq\frac{2}{I\eta}\left[{F}(\boldsymbol{\overline{\boldsymbol{\omega}}^{1}})-{F}(\boldsymbol{\overline{\boldsymbol{\omega}}^{*}})\right]+\\
   &\frac{1}{I}\sum_{i=1}^{I}
   \sum_{\forall k}\sum_{\forall n}\Big[\Gamma_{k,n}(R_1,R_2)\mathbb{E}(|{\mathcal{D}_n}|-\alpha_n^{t_i} |{\mathcal{D}_n}|)+
   \Psi_{k,n}(R_1,R_2)\Big],
\end{aligned} \label{upp}
\end{equation}}%
where $m_i=\lceil\frac{i}{R_1R_2}\rceil$, $t_i=\lceil\frac{i}{R_2}\rceil$, $\boldsymbol{\overline{\boldsymbol{\omega}}^{*}}$ is the optimal global model, $\Gamma_{k,n}(R_1,R_2)$ and $\Psi_{k,n}(R_1,R_2)$ are two increasing functions of $R_1$ and $R_2$. 
The convergence upper bound in~\eqref{upp} is obtained by extending the analysis presented in \cite[Theorem 2]{2022Adaptive}. In particular, persistent client association was considered throughout an HFL process in \cite[Theorem 2]{2022Adaptive}.
In this paper, we extrapolate that analysis by considering the situation in which the clients may be associated with different edge servers at different edge aggregation rounds.

The upper bound of the expected global gradient deviation on the right-hand side (RHS) of \eqref{upp} decreases with an increasing number of participating
clients, $|{\Omega ^t}|:=\sum_n \alpha _n^t$, and increases with the numbers of local iterations in edge and global aggregations, $R_2$ and $R_1R_2$.
Given $R$, $R_1$, and $R_2$, minimizing the upper bound 
% \on the RHS of \eqref{upp} 
can be transformed equivalently to maximize the number of participating clients $|{\Omega ^t}|$ in each round, since the upper bound decreases with the increasing number of participating clients.
%To accomplish an FL task with a reasonable learning delay and model accuracy, we optimize resource allocation and client schedule in this section. It is difficult to determine the exact accuracy of the global model until a test dataset is used to evaluate the outcomes. Nevertheless, empirical studies \cite{luo2020hfel,00925} and analysis \cite{9148862,liu2021hierarchical} have demonstrated more participating clients in an FL process can improve the convergence rate and model accuracy. 
On the other hand, scheduling more clients may incur a severer straggler effect and slow the FL process down,  especially when there are limited communication and computation resources. 
We set ${O_t} = \lambda |{\Omega ^t}| - {T^t}$ to be the objective function in the $t$-th edge aggregation round to jointly assess the model accuracy and learning delay of HFL, where 
% $|{\Omega ^t}|$ is the number of participating clients and 
$\lambda$ is a positive control parameter that balances the number of scheduled clients and learning delay. 

Since completing an FL task requires $R$ rounds of cloud aggregation, $RR_1$ rounds of edge aggregation, and $RR_1R_2$ rounds of local iterations, the overall learning delay is $\sum_{t=1}^{RR_1}T^t+RT_g$, where $T^t=\max_k\{{\max}_{n \in {\mathcal{Z}}_k^{t}} (\alpha _n^t\frac{{{R_2}{c_n}M\beta}}{{f_n^{t}}}+ \alpha _n^t\frac{\zeta }{{r_{nk}^{t}}}+T_e)\}$. %Considering an HFL process of $R$ cloud aggregation rounds, 
We aim to maximize the utility function ${U} := \sum_{t = 1}^{R{R_1}} {{O_t}}- R{T_g}$ by jointly optimizing the client selection $\alpha _n^t$, client association $\mathcal{Z}_k^t$, bandwidth allocation $b_{nk}^{t}$, and the CPU frequencies and transmit powers of the clients, $f_n^t$ and $p_n^{t, com}$. 
Since the objective is to maximize the number of scheduled clients and minimize the learning delay of an HFL process, the system tends to select clients in good channel conditions. As a consequence, some clients remaining in poor channel conditions may never be selected. However, these clients may have datasets critical for the model accuracy, especially when heterogeneous data is considered among the clients. To avoid this, we define $F$ to be the maximum number of edge aggregation rounds between two consecutive selections of a client, i.e., $\alpha _n^t=1$ if $t-{\tau_{n,t}} = F$, where 
${\tau_{n,t}} $ indicates the latest edge aggregation round when client $n$ was selected before the current $t$-th round.

%{\color{red} \textbf{COMMENT:} Please definite the notations more consistently. If $t$ indicates round, now $l$ also indicates round. That would be confusing to the reviewers. How about $\tau_{n,t}$?}

Let $\boldsymbol{A_t} = \{\alpha _n^t, f_n^t,p_n^{t,com}, {\mathcal{Z}}_k^t,b_{nk}^t \text{} | \text{ } n \in {\mathcal N}, k \in {\mathcal K}\}$ collect the optimization variables. The problem considered is
\begin{subequations}\label{P1}
\begin{align} 
\quad & \mathop {{\rm{max}}}\limits_{\boldsymbol{A_t}} \quad U= \sum\limits_{t = 1}^{R{R_1}} {{O_t}} - R{T_g} \label{Ta}\\
\text{s.t. } 
&  E_n^{t} + E_{n, ho}^{t} \geq E_n^{t, cmp}+ E_n^{t, com},\quad \forall n,t, \label{Tb} \\
&  E_{n, c}^{t} =\min \{  E_n^{t} + E_{n, ho}^{t} - E_n^{t, cmp} - E_n^{t, com}, E^{\max} \}, \label{Tb1}\\
&  E_n^{t+1} = \min \{ E_{n, c}^{t} + E_{n, hi}^{t}+E_{g}^{t}, {E^{\max}}\}, \label{Tb2}\\
& \alpha _n^t = \{ 0,1\}, \quad\forall n,t, \label{Tc}  \\
& \sum\limits_{n \in {{\mathcal{Z}}_k^t}} {{b_{nk}^{t}} = 1},~b_{nk}^{t}\in [0,1], \quad\forall k,t, \label{Te} \\
& \alpha _n^t=1,~\text{if}~t-{\tau_n^t} = F, \quad\forall n,t, \label{Tf} \\
& 0 \le p_n^{t, com} \le p_n^{\max }, \quad\forall t,n \in {\Omega^t}, \label{Tg} \\
&0 \le f_n^{t} \le f_n^{\max }, \quad\forall t,n \in {\Omega^t}, \label{Th} \\
& {\mathcal{Z}}_i^{t} \cap {\mathcal{Z}}_k^{t} = \emptyset,\quad \forall t,i \ne k,i,k \in {\mathcal K} . \label{Tk}
\end{align}
\end{subequations}
Constraints \eqref{Tb}--\eqref{Tb2} ensure the energy causality. \eqref{Tc} and \eqref{Te} provide the feasibility conditions of client selection and bandwidth allocation, respectively. \eqref{Tg} and \eqref{Th} specify the feasible regions of the CPU frequencies and transmit powers of the clients, respectively. \eqref{Tk} ensures each selected client is associated with only one edge server. 

As shown in (18b), energy harvesting affects the available energy in the battery $E_n^{t}$, and the latter further constrains the energy consumption of local computing $E_n^{t, cmp}(\alpha _n^t, p_n^{t, com})$ and local model uploading $E_n^{t, com}(\alpha _n^t, f_n^t)$ at the clients. As a consequence, the decisions of the client selection $\alpha _n^t$, as well as the CPU frequencies and transmit powers of the clients, $f_n^t$ and $p_n^{t, com}$, all depend on the energy harvesting. On the other hand, more participating clients (i.e., larger $|{\Omega ^t}|=\sum_n \alpha _n^t$) can help improve the model accuracy. Meanwhile, larger $f_n^t$ and $p_n^{t, com}$ can shorten the training delay, given the computation latency $T_n^{t, cmp} = \alpha _n^t\frac{{{R_2}{c_n}M\beta}}{{f_n^{t}}}$ and the communication latency $T_n^{t, com} = \alpha _n^t\frac{\zeta }{{r_{nk}^{t}}(p_n^{t, com})}$. To this end, energy harvesting, model accuracy, and training delay are highly interdependent.

It is challenging to solve problem \eqref{P1} directly. 
First, obtaining the optimal solution to problem \eqref{P1} would require the perfect \textit{a-priori} knowledge of the system (including channel conditions and energy arrival rates) throughout the HFL process, which is impossible in practice. 
Even if the \textit{a-priori} knowledge were available, problem \eqref{P1} would still be an intractable mixed-integer nonlinear programming problem (MINLP) that is typically NP-hard. 
Further, the evolution of the battery energy level in \eqref{Tb1} and \eqref{Tb2} leads to the coupling of the optimization variables over time. Specifically, the resource allocation and client scheduling in the past can affect the current battery energy level and, in turn, the current resource and client scheduling decisions. For example, if high CPU frequencies or transmit powers are nearsightedly decided to maximize the current $O_t$ in the $t$-th edge aggregation round, there is no other option but to take low CPU frequencies and transmit powers for a lower $O_{t+1}$ in the ($t+1$)-th edge aggregation round due to insufficient energy in the battery.

\section{Proposed TP-DDPG Scheme}
As problem \eqref{P1} entails a Markov decision process (MDP), we resort to DDPG, the state-of-the-art DRL technique for learning effective decisions in complex and dynamic environments. However, direct use of DDPG to solve problem \eqref{P1} may not converge, since, as the numbers of clients and edge servers grow, the MDP is increasingly complex with the number of variables in $\boldsymbol{A_t}$.
In this section, we delineate the new TP-DDPG framework to solve problem \eqref{P1} with a significantly improved convergence rate.
In the first phase, the selection of participating clients, transmit powers, and CPU configurations are decided by the DDPG for each edge aggregation round. In the second phase, the rest of problem \eqref{P1} is efficiently solved using the new SCABA, which is interpreted as part of the environment for the DDPG and produces rewards for the DDPG agent for model training.
%{\color{blue}Although DDPG is suitable for learning effective decisions in complex dynamic environments, it requires a considerable number of iterations for training and may even not converge due to the randomness-based exploration procedure over a large action space \cite{ye2019drag,kim2020reinforcement}. With reduced action space, the number of decisions the DDPG agent needs to make is substantially reduced, leading to faster convergence and enhanced stability.}

\begin{figure*}
    \centering
    \includegraphics[scale=0.8]{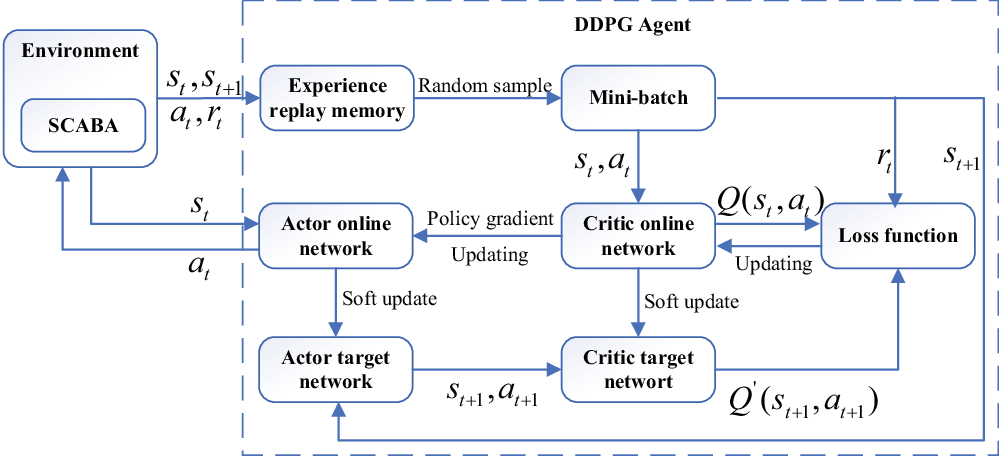}
    \caption{The structure of the proposed TP-DDPG algorithm, where a DDPG agent consisting of actor-critic networks and an experience replay memory makes decisions on client selection, power control and CPU configuration, and the SCABA decides on client association and bandwidth allocation. The SCABA can be interpreted as part of the environment for the DDPG, and produces a reward for each decision that the DDPG agent makes on client selection, transmit power, and CPU configuration.}
    \label{fig:DDPG_stucture}
\end{figure*}

\subsection{Discrete-Time MDP Framework for RL}

Being an MDP, problem (\ref{P1}) can be characterized using a 3-tuple $(\boldsymbol {S}, \boldsymbol {A}, \boldsymbol {r})$, where $\boldsymbol {S}$, $\boldsymbol {A}$, and $\boldsymbol {r}$ are the state, action, and reward of the MDP, respectively.

\subsubsection{\textbf{State $\boldsymbol {S}$}}
In the $t$-th edge aggregation round, the system state $\boldsymbol {s}_t\in \boldsymbol {S}$ is defined to be $\boldsymbol {s}_t = \{ E_{n, c}^{t-1},E_{n}^{t},h_{nk}^{t},\tau_n^t,\forall n \in {{\mathcal N}},\forall k \in {{\mathcal K}}\}$. 
Recall that $E_{n, c}^{t-1}$ is the battery energy level of client $n$ at the end of the \textit{on} time in the $(t-1)$-th edge aggregation round; $E_{n}^{t}$ is the battery charging level of client $n$ when the $t$-th edge aggregation round begins; $h_{nk}^{t}$ is the channel gain from client $n$ to edge server $k$ in  the $t$-th edge aggregation round; and ${\tau_n^t}$ indicates the latest edge aggregation round when client $n$ was selected before the current $t$-th round.

\subsubsection{\textbf{Action $\boldsymbol {A}$}}

To reduce the action space and accelerate convergence, we design the action of the DDPG model in the $t$-th edge aggregation round to be $\boldsymbol {a}_t = \{\alpha _n^t, f_n^{t},p_n^{t,com},\forall n\}\in \boldsymbol {A}$. Here, $\alpha _n^t$ is the binary client selection decision; $f_n^{t}$ and $p_n^{t,com}$ are the CPU frequency and transmit power of client $n$, respectively, if the client is selected. 
%The DDPG can learn to efficiently utilize the energy in the clients' batteries by optimizing $\boldsymbol {a}_t$.

\subsubsection{\textbf{Reward $\boldsymbol{r}$}}
A reward evaluates the action executed. For problem \eqref{P1}, a straightforward design principle would set the instantaneous objective function in \eqref{Ta}, i.e., $O_t$, as the immediate reward. In contrast, we define the reward to be an exponential function of the instantaneous objective, i.e., 
\begin{equation}\label{instantaneous reward}
r_t = {e^{(c+O_t)}}-\varphi,   
\end{equation}
where $c$ is a tunable parameter, and $\varphi$ is the penalty if constraints (\ref{Tb}) and (\ref{Tf}) are unsatisfied. The immediate reward \eqref{instantaneous reward} is produced by a new straggler-aware client association and bandwidth allocation algorithm, given the action $\boldsymbol {a}_t = \{\alpha _n^t, f_n^{t},p_n^{t,com},\forall n\}$ taken by the DDPG.

\subsection{Proposed DDPG-Based Client Selection, Power Control and CPU Configuration}

At any edge aggregation round $t$, the DDPG agent deployed at the cloud server perceives the state  ${\boldsymbol{s}_t}$ and executes action ${\boldsymbol {a}_t}$. 
After executing the action, the environment feeds back a scalar reward ${r_t}$ and transfers from state ${\boldsymbol{s}_t}$ to ${\boldsymbol{s}_{t + 1}}$. Let $\pi (\boldsymbol{s}_t)$ denote the policy that projects state ${\boldsymbol{s}_t}$ to action ${\boldsymbol{a}_t}$. Let ${Q^\pi }(\boldsymbol{s},\boldsymbol {a})$ denote the action-value function that represents the expected accumulative discounted reward over infinite time under the policy $\pi$ with the initial state $\boldsymbol{s}$ and the initial action $\boldsymbol {a}$, as given by:
\begin{align}
    {Q^\pi }(\boldsymbol {s},\boldsymbol {a}) = {\mathbb{E}_\pi }\bigg [\sum\limits_{i  = 0}^\infty  {{\gamma ^i }r_{t + i}|{\boldsymbol {s}_t} = \boldsymbol {s},{\boldsymbol{a}_t} = \boldsymbol {a}} \bigg],
\end{align}
where $\gamma \in [0,1]$ denotes the discount factor. 

The agent aims to learn the optimal policy ${{\pi ^*}(\boldsymbol {s})}$:
\begin{equation}
    {\pi ^*}(\boldsymbol {s}) = \mathop {\arg \max }\limits_{\boldsymbol {a}} {Q^*}(\boldsymbol {s}, \boldsymbol {a}),
\end{equation}
where $Q^*{(\boldsymbol {s}, \boldsymbol {a})} = \mathop {\max }\limits_\pi  {Q^\pi }(\boldsymbol {s}, \boldsymbol{a})$ is the optimal action-value function.
Among many DRL algorithms, DDPG is suited for our considered problem because of its capability to cope with the continuous state and action spaces. With the actor-critic framework, the DDPG model applies the actor-network to fit the policy $\pi$, and the critic network to approximate the action-value function ${Q^\pi }(\boldsymbol {s}, \boldsymbol {a})$. An actor network or critic network contains two sub-networks with the same architecture: a target network and an online network. This ensures learning stability and prevents overestimation in large-scale problems.

Let ${\pi}({\boldsymbol {s}_t}|{\boldsymbol{\theta} ^{\pi}})$ and ${Q}({\boldsymbol {s}_t},{\boldsymbol {a}_t}|{\boldsymbol{\theta} ^{{Q}}})$ denote the actor and critic online networks, respectively; ${\boldsymbol{\theta} ^\pi }$ and ${\boldsymbol{\theta} ^Q}$ are the model parameters of the two deep neural networks (DNNs). Let ${\pi'}({\boldsymbol {s}_t}|{\boldsymbol{\theta} ^{\pi'}})$ and ${Q'}({\boldsymbol {s}_t},{\boldsymbol {a}_t}|{\boldsymbol{\theta} ^{{Q'}}})$ denote the actor and critic target networks, respectively; ${\boldsymbol{\theta} ^{\pi'}}$ and ${\boldsymbol{\theta} ^{Q'}}$ are the model parameters of the two DNNs. In order to update the critic online network, the following loss function is minimized:
\begin{align}
    L({\boldsymbol{\theta} ^Q}) = \frac{1}{{{M'}}}\sum\limits_t {[{{({y_t}-Q({\boldsymbol {s}_t},\pi ({\boldsymbol {s}_t|\boldsymbol{\theta}^ \pi})|{\boldsymbol{\theta} ^Q}) )}^2}]},
\label{onlinec}
\end{align}
where ${y_t} = r_t + \gamma Q'({\boldsymbol {s}_{t+1}},\pi '({\boldsymbol {s}_{t + 1}|\boldsymbol{\theta} ^{\pi '}})|{\boldsymbol{\theta} ^{Q'}})$ and $M'$ is the size of a mini-batch. We optimize the actor online network in the direction of ${\nabla _{{\boldsymbol{\theta} ^\pi }}}J({\boldsymbol{\theta} ^\pi }) \approx \frac{1}{M'}\sum\limits_t {{\nabla _{{\boldsymbol {a}_t}}}Q({\boldsymbol {s}_t},{\boldsymbol {a}_t}|{\boldsymbol{\theta} ^{\rm{Q}}})} {\nabla _{{\boldsymbol{\theta} ^\pi }}}\pi ({\boldsymbol {s}_t}|{\boldsymbol{\theta} ^\pi })
$
to maximize the following policy objective function:
\begin{align}
J({\boldsymbol{\theta} ^\pi }) = {\mathbb{E}_{{\boldsymbol{\theta} ^\pi }}}[Q({ \boldsymbol {s}_t},\pi ({\boldsymbol {s}_t}|{\boldsymbol{\theta} ^\pi })|{\boldsymbol{\theta} ^Q})],
\label{J}
\end{align}
where ${\nabla _{{\boldsymbol{\theta} ^\pi }}}$ denotes the derivative w.r.t $\boldsymbol{\theta} ^\pi$. 
Then, the actor and critic target networks are updated softly by
\begin{align}
&{\boldsymbol{\theta} ^{{\pi '}}} \leftarrow \phi {\boldsymbol{\theta} ^\pi } + (1 - \phi ){\boldsymbol{\theta} ^{{\pi'}}}; \label{targeta}\\
&{\boldsymbol{\theta} ^{{Q'}}} \leftarrow \phi {\boldsymbol{\theta} ^Q} + (1 - \phi ){\boldsymbol{\theta} ^{{Q'}}}, \label{targetc} 
\end{align}
where $\phi$ is a parameter controlling the learning speed.

The proposed TP-DDPG algorithm is depicted in Fig.~\ref{fig:DDPG_stucture}, and summarized in Algorithm 1. In each edge aggregation round, the DDPG agent collects necessary state information $\boldsymbol {s}_t$ from the environment. The clients and the edge servers upload their local observations, together with their trained models. The cloud server sends back its decisions on the actions, along with the global model. The latency and energy consumption of transmitting the states and actions are comparatively negligible, since their size is negligible compared to the model size. %We can ignore the information collection delay due to the comparatively negligible size of the state information. 
The actor online network outputs action $\boldsymbol {a}_t$. Given $\boldsymbol {a}_t$, we optimize the remaining decisions $\{{\mathcal{Z}}_k^t,b_{nk}^t, \forall n, k\}$ to evaluate the reward, as will be articulated in Section V-C. 
The agent obtains the reward $r_t$ at the end of edge aggregation $t$ and observes a new state $\boldsymbol {s}_{t+1}$.

%There are two stages of the DDPG-based solution: training and inferring. In the training stage, 
An experience replay pool is added to the DDPG agent to preserve the experience $\{ {\boldsymbol {s}_t},{\boldsymbol {a}_t},{r_t},{\boldsymbol {s}_{t + 1}}\}$ in each iteration. The learning process commences when the experience replay buffer is full. Specifically, a mini-batch of $M'$ experiences is obtained by randomly sampling the replay buffer to train the DDPG network. The critic and actor online networks are then updated by minimizing the loss function in \eqref{onlinec} and maximizing the policy objective function in \eqref{J}, respectively, followed by the update of the target networks via \eqref{targeta} and \eqref{targetc}. 
The model can converge after hundreds of episodes by suppressing the correlations between observations and exploring different environment states. %Historical information of channel gains and energy arrivals is used to train the DDPG in an offline manner. In the inferring stage, the trained actor network can generate its action based on current environment states in real time.

\begin{algorithm}  
\SetKwRepeat{Do}{do}{while}
Initialize the actor online network and critic online network with random model parameters ${\boldsymbol{\theta} ^\pi }$ and ${\boldsymbol{\theta} ^Q}$, respectively.\\
Initialize the actor target network and critic target network with model parameters ${\boldsymbol{\theta} ^{{\pi'}}} \leftarrow {\boldsymbol{\theta} ^ \pi}$ and ${\boldsymbol{\theta} ^{{Q'}}} \leftarrow {\boldsymbol{\theta} ^Q}$, respectively.\\

Initialize the experience replay buffer with size $V$.\\
Parameter updating:\\
\For{each episode}{{Initialize the environment and receive the initial observed system state $\boldsymbol {s}_1$.\\
\For{ edge aggregation round $t=1, 2, \ldots, R R_1$}{
Choose action ${\boldsymbol{a}_t} = \pi ({{\bf{s}}_t}|{{\bf{\theta }}^\pi }) + {N_0}$ via the actor online network with the exploration noise ${N_0}$.\\
 Given ${\boldsymbol {a}_t}$, obtain the client association and bandwidth allocation strategy by using Algorithm 2. \\
 Obtain reward $r_t$ and the subsequent state $\boldsymbol {\mathop s}\nolimits_{t + 1}$. \\
Save experience $\left\{ {\boldsymbol {\mathop s}\nolimits_t ,\boldsymbol {\mathop a}\nolimits_t ,\mathop r\nolimits_t ,\boldsymbol {\mathop s}\nolimits_{t + 1} } \right\}$\ in the replay buffer. \\
 \If{the experience replay buffer is full}{Randomly sample $M'$ transitions from the buffer and input them to the actor and critic networks. \\
Update the critic and actor online networks by  minimizing \eqref{onlinec} and maximizing \eqref{J}, respectively. \\
Update the target networks via \eqref{targeta} and \eqref{targetc}.}
 }  }    
  }
\caption{Proposed TP-DDPG Algorithm}
\end{algorithm}

\subsection{Straggler-Aware Client Association and Bandwidth Allocation }

To mitigate the straggler effect in an edge aggregation round for the considered synchronous FL updates, we propose to use DDPG in the first phase of TP-DDPG to adaptively select clients preferably in good channel conditions in a round and avoid selecting straggling clients. Then, we propose the SCABA in the second phase to identify the
straggling edge server in each iteration, hence reducing its latency for edge aggregation. This is done by iteratively adjusting the association and bandwidth allocation strategy until the latency cannot be further shortened. In this way, the straggler effect in a global aggregation round can be significantly alleviated.

Given the action $\boldsymbol {a}_t = \{\alpha _n^t, f_n^{t},p_n^{t,com},\forall n\}$ taken by the proposed DDPG in the $t$-th edge aggregation round, problem \eqref{P1} is reduced to: 
\begin{align}
\mathop {\min }\limits_{{\mathcal{Z}}_k^{t}, b_{nk}^{t}} T^t  \quad \text{s.t.} \quad  \eqref{Tb},\,  \eqref{Te}, \, \eqref{Tk},  \label{Ua}
\end{align}
where the number of participating clients $|\Omega^t|$ is suppressed from the original objective function, i.e., $|\Omega^t|= \sum_n \alpha _n^t$.

The client association and bandwidth allocation problem in \eqref{Ua} is to distribute the clients among the edge servers and allocate the bandwidth of the edge servers to the selected clients. The problem is a complex combinatorial optimization problem.
%Due to the nature of edge association and bandwidth allocation problem, solving the optimal value of problem (\ref{Ua}) is very difficult.} 
%In particular, the client association per edge aggregation round could be obtained by exhaustive search. However, it is computationally prohibitive when the numbers of clients and edge servers are large. An alternative solution is associating a client with the edge server with the strongest channel gain. Yet, its performance is poor when the clients are non-uniformly distributed, and some edge servers are not associated with any client.
%
% In what follows, we propose a heuristic client association and bandwidth allocation algorithm  with a low complexity to balance the trade-off between delay and performance. 
%
In what follows, we decompose problem \eqref{Ua} into client association and bandwidth allocation subproblems, and delineate the SCABA, which balances learning delay and accuracy with low complexity.

First, the SCABA initializes the client association decision ${{\mathcal{Z}}^{t}} = \{{\mathcal{Z}}{_k^{t}}:k \in {\mathcal K}\}$ by connecting each selected client to the edge server with the strongest channel gain. Given ${\mathcal{Z}_k^t}$, the bandwidth allocation sub-problem concerning edge server $k$ can be written as
\begin{subequations}\label{Va}
\begin{align}
\quad & \quad 
\mathop  {\min }\limits_{b_{nk}^t}~T_k^t  \\ \quad\text{s.t.} &\quad  \sum\limits_{n \in {{\mathcal{Z}}_k^t}} {{b_{nk}^{t}} = 1},~b_{nk}^{t}\in [0,1], \quad\forall k, t. 
\end{align}
\end{subequations}
Recall $T_k^{t} = \mathop {\max}\limits_{n \in {\mathcal{Z}}_k^{t}} (T_n^{t, cmp} + T_n^{t, com}+T_e)$. By introducing an auxiliary variable $T$, the problem in \eqref{Va} is rewritten as
\begin{subequations}\label{Vb}
\begin{align}
\quad & \quad  \mathop  {\min }\limits_{b_{nk}^t}~T  \\
\quad\text{s.t.} & \quad T \geq \mathop{T_n^{t,cmp} + T_n^{t, com}+T_e}, \quad\forall {n \in {\mathcal{Z}}_k^{t}}. \label{27b} \\
 &\quad  \sum\limits_{n \in {{\mathcal{Z}}_k^t}} {{b_{nk}^{t}} = 1},~b_{nk}^{t}\in [0,1], \quad\forall k,t.\label{27c} 
\end{align}
\end{subequations}
Given the action $\boldsymbol {a}_t = \{\alpha_n^t, f_n^{t},p_n^{t,com},\forall n\}$, $T_n^{t,cmp}$ and $T_e$ are constants, and $T_n^{t, com}$ is inversely proportional to $b_{nk}^t$. Since the objective function and the inequality constraint \eqref{27b} are convex while the equality constraint \eqref{27c} is an affine function, problem \eqref{Vb} is convex and can be solved optimally using convex solvers. We use the off-the-shelf function \textit{fminimax} in Matlab toolkit to solve problem \eqref{Va}. %Given ${\mathcal{Z}_k^t}$, the bandwidth allocation sub-problem concerning edge server $k$ can be written as:
% \begin{align}\label{Va}
% \mathop  {\min }\limits_{b_{nk}^t}~T_k^t \quad\text{s.t.} \quad \eqref{Te}. 
% \end{align}
%Problem (\ref{Va}) can be solved optimally by the off-the-shelf function \textit{fminimax} in Matlab toolkit, as the \textit{fminimax} function can reformulate such a min-max problem into a convex minimization problem due to convexity and linearity in its objective and constraints, respectively.
After solving problem~(\ref{Va}) for all edge servers for $k \in {\mathcal K}$, we can obtain the minimum edge aggregation delay $T^t=\mathop {\max }\limits_{k \in {\mathcal K}} T_k^{t}$.

Next, we update the client association $\mathcal{Z}^t$ by moving or swapping the clients associated with the straggler, and then solve problem \eqref{Va} again until the edge aggregation delay $T^t$ cannot be further reduced. Particularly, we propose to update the client association in the following ways. 
\begin{enumerate}
\item We can obtain a new client association strategy $\mathcal{Z}^t$ by removing client $n$ from  $\mathcal{Z}_{v^*}^{t}$, the set of clients associated with the straggler $v^*$, to another set $\mathcal{Z}_l^{t}$;
\item Or we can pick one client $n \in \mathcal{Z}_{v^*}^{t}$ and one client $n' \in \mathcal{Z}_l^{t}$, and swap their association strategy. 
\end{enumerate}
A historical straggler set, denoted by $\mathcal{H}$, is maintained in every edge aggregation round $t$ to record the straggling edge server $k$ and its associated clients $\mathcal{Z}_k^t$. If a server-client pair $(k,\mathcal{Z}_k^t)$ in a new client association $\mathcal{Z}^t$ satisfies  $(k,\mathcal{Z}_k^t) \in \mathcal{H},~\exists k \in \mathcal{K}$, we skip $\mathcal{Z}^t$ since its corresponding edge aggregation delay cannot be shorter than the minimum delay under the explored client associations. A straggler is the last edge server required to complete the edge aggregation. It is the bottleneck of edge aggregation. If a client association has caused a straggler problem, it would not be assessed again.

The key idea of the SCABA is to identify the straggling edge server in each iteration, and reduce its latency for completing edge aggregation by repeatedly adjusting its association and bandwidth allocation strategy until the latency cannot be further shortened. Algorithm~2 summarizes the proposed SCABA, where ${{\mathcal{B}}^{t}} = \{ {b_{nk}^{t}}| n \in {{\Omega} ^{t}},k \in {\mathcal K}\}$, and ${\tilde{\mathcal{T}}^t} = \{ T_k^t|k \in \mathcal K\}$. 
We first initialize ${{\mathcal{Z}}^{t*}}$ by connecting each selected client to the edge server with the strongest channel gain, and obtain ${{\mathcal{B}}^{t*}}$ and ${\tilde{\mathcal{T}}^{t*}}$ by solving problem (\ref{Va}) in Steps~1 and 2. Next, we find the straggling edge server $v^*$. Then, we iteratively adjust the client association (by switching clients away from the straggler server or swapping the clients with those associated with other servers) to minimize the learning period of the straggler until the delay cannot be shortened or the maximum number of iterations, denoted by $\xi$, is reached. 
%adjustment attempts. $\xi$ is a predefined value. The adjustments include: 
% \begin{enumerate}
% \item In Step~6, we obtain a new client association strategy $\mathcal{Z}^t$ by removing client $n$ from the set $\mathcal{Z}_{v^*}^{t}$ that collects the clients associated with the straggler $v^*$, and then adding this client to another set $\mathcal{Z}_l^{t}$;
% \item In Step~17, we pick one client $n \in \mathcal{Z}_{v^*}^{t}$ and one client $n' \in \mathcal{Z}_l^{t}$, and exchange their association strategy. 
% \end{enumerate}

In each iteration, only when the edge aggregation delay $T_v^{t}$ corresponding to the client association ${\mathcal{Z}}^t$ and bandwidth allocation ${\mathcal{B}^t}$ is shorter than  ${T_{v^*}^{t*}}$, should the optimal client association and bandwidth allocation decisions be updated, together with the delay and straggler; see Steps~10--13.
By switching out or swapping the clients associated with the straggling edge server to shorten the edge aggregation delay, the SCABA can converge to a stable and optimal client association and bandwidth allocation strategy within a limited number of iterations, where each edge server owns a stable set of associated clients to achieve the minimum edge aggregation delay. In other words, the delay cannot be further shortened by further adjusting the client association strategy.

%{\color{red}\textbf{COMMENT:} This paragraph is difficult to understand. Are you basically saying if a client association has already been known to cause a straggler problem, there is no point in assessing the association again? Also, the key question is how to perturb the association.}

\subsection{Analysis of Computational Complexity}

% The time complexity is defined as the running time needed by an algorithm, which can be defined as a function of the size of input parameters. The running time depends on the number of operations. Because it is hard to determine the exact number of operations, the big $\mathcal{O}$ notation is employed to obtain an asymptotic upper bound of the number of operations. 
In the first phase of the proposed TP-DDPG framework, the DDPG-based algorithm generates action ${\boldsymbol {a}_t}$. In DNNs, the complexity is dependent on the specification.
Suppose that the actor and critic networks have $I_a$ and $I_c$ fully connected layers, respectively. The complexity of the DDPG-based algorithm is $\mathcal{O}(\sum\limits_{i = 0}^{{I_a} - 1} {v_i^av_{i + 1}^a}  + \sum\limits_{i = 0}^{{I_c} - 1} {v_i^{\rm{c}}v_{i + 1}^c})$, where $v_i^a$ and $v_i^c$ denote the numbers of neurons in the $i$-th layer of the actor and critic networks, respectively \cite{qiu2019deep}. 

In the second phase, the SCABA produces the client association and
bandwidth allocation with the complexity dominated by iteratively solving (\ref{Va}) under different client association policies.
Function \textit{fminmax} used to solve (\ref{Va}) for edge server $k$ employs a sequential quadratic programming (SQP) method, incurring the complexity of $\mathcal{O}(|{\mathcal{Z}}_k^t|^3)$ \cite{ghaemi2010robust}. Hence, solving (\ref{Va}) for all edge servers ($k=1,\cdots, K$) incurs the complexity of $\mathcal{O}(|\Omega^t|^3)$ in the worst-case scenario. Let $G$ denote the number of attempts to adjust the client association. 
%(by randomly moving a client associated with the straggler to {\color{blue}another edge server or exchanging the association of two clients under two edge servers}. 
The computational complexity of the SCABA is $\mathcal{O}(G|\Omega^t|^3)$.

The proposed TP-DDPG algorithm offers the advantage of both offline training and online testing. Offline training boasts computational efficiency, quicker convergence, and optimal utilization of available data. Online testing enables the agent to adapt and refine its policy in real-time. Specifically, online testing facilitates continuous improvement of the policy by evaluating its performance and making adjustments on-the-fly based on real-time feedback \cite{shuyan2024}.

%Considering $|Z_k^t|$ is different among all edge servers and the total selected clients is ${|\Omega|}^t$, we assume the time complexity of solving problem (\ref{Va}) is $\mathcal{O}({|\Omega{^t}|}^{3})$ for simplicity. In step~8, in order to obtain ${{\mathcal{B}}^{t}}$ and ${\tilde{\mathcal{T}}^{t}}$, we only need to solve problem (\ref{Va}) given $\mathcal{Z}_{v*}^t$ and $\mathcal{Z}_l^t$, so the corresponding time complexity is $\mathcal{O}(|\mathcal{K}||\Omega{^t}|^3)$ from step~5 to step~15.  Similarly, the time complexity is $\mathcal{O}(|\mathcal{K}||\Omega{^t}|^3)$from step~16 to step~19. 
%According to these, the time complexity of the SCABA algorithm is $\mathcal{O}(I|\mathcal{K}||\Omega{^t}|^3)$, where $I$ is the number of iterations to break repeat in step 4. 

%As a result, the computational complexity of the TP-DDPG algorithm is $\mathcal{O}(\sum\limits_{i = 0}^{{I_a} - 1} {v_i^av_{i + 1}^a}  + \sum\limits_{i = 0}^{{I_c} - 1} {v_i^{\rm{c}}v_{i + 1}^c}+G|\Omega{^t}|^3)$.
%{\color{red} \textbf{COMMENT:} First, I am unsure if it is appropriate to add the two parts of complexity as such. The complexity of the DDPG is constant, i.e., $\mathcal{O}(1)$, in the context of the SCABA. Second, the complexity is just one iteration. How many DDPG iterations are required? DDPG only conducts the inference once until the clients move away/around? Maybe we just do not have the last paragraph. }

\begin{algorithm}
 \caption{Proposed SCABA}
 \KwIn{${{\Omega} ^{t}}$, $f_n^{t}$, $p_n^{t}$, $h_{nk}^{t}$, and ${\mathcal{H}}= \emptyset $, $\forall n \in {{\Omega}^{t}}$, $k \in  {\mathcal K}$.}
 \KwOut{Optimal client association ${{\mathcal{Z}}^{t*}}$, and bandwidth allocation ${{\mathcal{B}}^{t*}}$.}
Connect each selected client $n \in {{\Omega} ^{t}}$ to the edge server with the strongest channel gain to initialize ${{\mathcal{Z}}^{t*}}$. \\
Given ${{\mathcal{Z}}^{t*}}$, obtain ${{\mathcal{B}}^{t*}}$ and ${\tilde{\mathcal{T}}^{t*}}$ by solving (\ref{Va}).\\
Find the straggling edge server $v^* = \mathop {\arg \max }_{k \in  {\mathcal {K}}} {\tilde{\mathcal{T}}^{t*}}$, and put $(v^*,\mathcal{Z}_{v^*}^{t*})$ into $\mathcal{H}$. \\
\Repeat{No shorter delay can be obtained or $\xi$ times of adjustment attempts is reached}{

\For{$l \in {\mathcal {K}}, l \ne v^*$}{
Set $\mathcal{Z}^t=\mathcal{Z}^{t*}$, then randomly pick client $n \in {{\mathcal{Z}}_{v*}^{t}}$, and transfer client $n$ to ${{\mathcal{Z}}_l^{t}}$. \\
\If{$(k,\mathcal{Z}_k^{t}) \notin {\mathcal{H}},\forall k \in  {\mathcal K} $}{
Given ${{\mathcal{Z}}^{t}}$, obtain ${{\mathcal{B}}^{t}}$ and ${\tilde{\mathcal{T}}^{t}}$ by solving (\ref{Va}).\\
Set $v = \mathop {\arg \max }\limits_{k \in  {\mathcal {K}}} {\tilde{\mathcal{T}}^{t}}$, and put $(v,\mathcal{Z}_v^t)$ into $\mathcal{H}$.\\
\If{${T_{v^*}^{t*}}> T_{v}^{t}$}
{Set ${{\mathcal{Z}}^{t*}}={\mathcal{Z}}^{t}$, ${\mathcal{B}^{t*}}=\mathcal{B}^{t}$, ${\tilde{\mathcal{T}}^{t*}} = \tilde{\mathcal{T}}^{t}$ and $v^* = v$. \\
\textbf{break}.}
}}

\For{$l \in {\mathcal{K}}, l \ne v^*$}{
Set $\mathcal{Z}^t=\mathcal{Z}^{t*}$, then randomly pick client $n \in {{\mathcal{Z}}_{v*}^{t}}$ and $n' \in {{\mathcal{Z}}_l^{t}}$, and swap their association strategy, i.e., let $n \in {{\mathcal{Z}}_l^{t}}$ and $n' \in {{\mathcal{Z}}_{v*}^{t}}$.} 
Perform Steps 7-15.\\}
Return ${{\mathcal{Z}}^{t*}}$ and ${\mathcal{B}^{t*}}$. \\
 \end{algorithm}
\label{sec:prob}

\section{Performance Evaluation}
Performance evaluation is provided to demonstrate the effectiveness and merits of our proposed algorithm. %First, we describe the general setting of our simulation environment. Then, extensive numerical evaluations are performed to demonstrate the efficiency of our proposed algorithm. 

\subsection{Experimental Settings}
Consider an HFL system with three edge servers and ten clients uniformly distributed in a circular area with a radius of 250 m. A cloud server situates at the center of the area. 
Suppose that the channel gain between an edge server and a client yields the path loss model %yields the Rayleigh distribution with the mean ${10^{ - L(d)/10}}$, where 
$30\log (d) + 72.4$ (in dB) and Rayleigh fading, where $d$ (in km) is the distance between the edge server and the client \cite{yin2020joint}. The channel gain remains unchanged throughout a round of edge aggregation and changes independently between rounds. 
Suppose that a Poisson distribution governs the renewable energy arrival rate at each client, with the mean uniformly distributed within $[200,1000]$ mJ. The energy arrival rates change every second.
%The channel gain between an edge server and a client yields the Rayleigh distribution with the mean ${10^{ - L(d)/10}}$, where $L(d) = 20\lg (d) + 20\lg(2450) + 32.45$ and $d$ (in kilometers) is the distance between the edge server and the client. Suppose that a Poisson distribution governs the renewable energy arrival rate at each client, with the mean uniformly distributed within the range of $[200,1000]$ mJ. 
By rigorously testing different values and assessing their corresponding convergence and performance, we set $c=5$ and $\varphi=5000$ for the reward function in \eqref{instantaneous reward}.

Consider classification FL tasks of handwritten digits using the MNIST dataset with 10 digit labels $\{0,1,\cdots, 9\}$ and a convolutional neural network (CNN) model. Each client is assigned 2000 heterogeneous training samples with two labels, e.g., client 1 owns labels $\{0, 2\}$ and client 2 owns labels $\{6, 7\}$.
%{\color{red} \textbf{COMMENT:} So the other data in the MNIST is not used? If this is the case, don't mention $\{0,1,\cdots, 9\}$ above!} 
The total number of cloud aggregations is $R=150$. In each cloud aggregation, the number of edge aggregation and local updates are 5 and 100, respectively. The other parameter settings are provided in Table \ref{tab:first}. 

\begin{table}[h]
\renewcommand{\arraystretch}{1.2}
\centering
	\caption{Parameters for performance analysis}
\label{tab:first}
	\begin{tabular}{p{5cm}|p{2cm}}
\hline  
\textbf{Parameter}&\textbf{Value}\\
\hline
Number of edge servers, $K$ & 3\\
Number of clients, $N$ & 10\\
Bandwidth of an edge server, $B$ & 1 MHz\\

Client transmission power, $p_n^{t, com}$  & [0, 1] W\\

Client CPU frequency, $f_n^{t}$  & [0, 3] GHz\\

CPU cycles computing per 
bit data, $c_n$ & [30, 100] \\

Background noise, $\psi$ & $10^{-9}$ W\\

%CNN learning rate & $0.00005$ \\

Uploaded model size, $\zeta$ & 0.2 MB\\

Batch size, $M$ and $M'$ & 32 \\

Effective capacitance coefficient, ${u_n}$ & $2\times10^{-28}$  \\

Discount factor, $\gamma$ & 0.99\\
The maximum round for not selecting a client, $F$ & 3\\
Times of adjustment attempts, $\xi$ & 5\\
Dataset & MNIST\\
DDPG memory size & 40000 \\

DDPG actor learning rate & 0.0001 \\

DDPG critic learning rate &0.0002 \\

\hline
\end{tabular}
\end{table}
%In order to evaluate the performance of the proposed algorithm, we present the following algorithms for comparison purpose.
No existing works has considered a holistic optimization of client scheduling, bandwidth allocation, and the CPU frequencies and transmission powers of the clients in an energy-harvesting HFL system. With due diligence, we compare the proposed TP-DDPG algorithm against the baselines listed below.
% 1) ALL clients selected (ACS): All clients participate in each edge aggregation of FL. 
% The optimization method is basically the same as our proposed algorithm, except that the action space and reward are changed to $A = \{ f_n^q,p_n^q\} ,n \in N$ and $R = {d^{(c - max(T_k^{q\_edge}) )}}$.
\begin{enumerate}
 \item Greedy Association (GA): Client $n$ is associated with the edge server with the strongest channel gain. 

 \item Even Bandwidth  Allocation (EBA): The bandwidth of edge server $k$ is evenly allocated to its associated clients.

% 4) Random CPU frequency and transmission power (RCFTP):  The cpu frequency and transmission power of each selected client are randomly determined within the allowable range. 

 \item Random Selection (RS): Each edge server randomly selects a fixed number of clients in an edge aggregation round. The number of selected clients is configurable.

% 6) Priority strategy (PS): Sort all clients according to battery energy, and then select a fixed number of clients with more battery energy.

% 7) Greedy client scheduling(GCS):  Select as many clients as possible within a given deadline.

 \item Negative Strategy (NS): The decision variable $\boldsymbol{A_t}$ remains unchanged during a cloud aggregation and only varies between different cloud aggregations.
 \item DDPG-Only: All decisions in $\boldsymbol{A_t}$ are generated using only the DDPG. 
 \item Holistic Optimization (HO): In each round of edge aggregation, a fixed number of clients with the largest energy stored in their batteries are selected and associated with the edge server with the strongest channel gain. Subsequently, the bandwidth allocation, CPU frequency, and transmission power decisions are holistically optimized locally through the off-the-shelf function \textit{fminimax} in Matlab toolkit.
  \item Multi-agent deep deterministic policy gradient (MADDPG): To compare the proposed TP-DDPG algorithm with its decentralized counterpart, we design and test the MADDPG, where a DDPG agent is placed at each client, enabling the clients to generate resource allocation and client scheduling decisions in parallel. (No agent can be placed at an individual edge server, since the clients associated with a server can change dynamically.) After each client/agent obtains the client schedule, CPU frequency and transmission power, the edge servers decide the bandwidth allocation under the constraint of the total available bandwidth of an edge server.
 \end{enumerate}

\subsection{Numerical Results}

\begin{figure}
    \centering
    \includegraphics[scale=0.5]{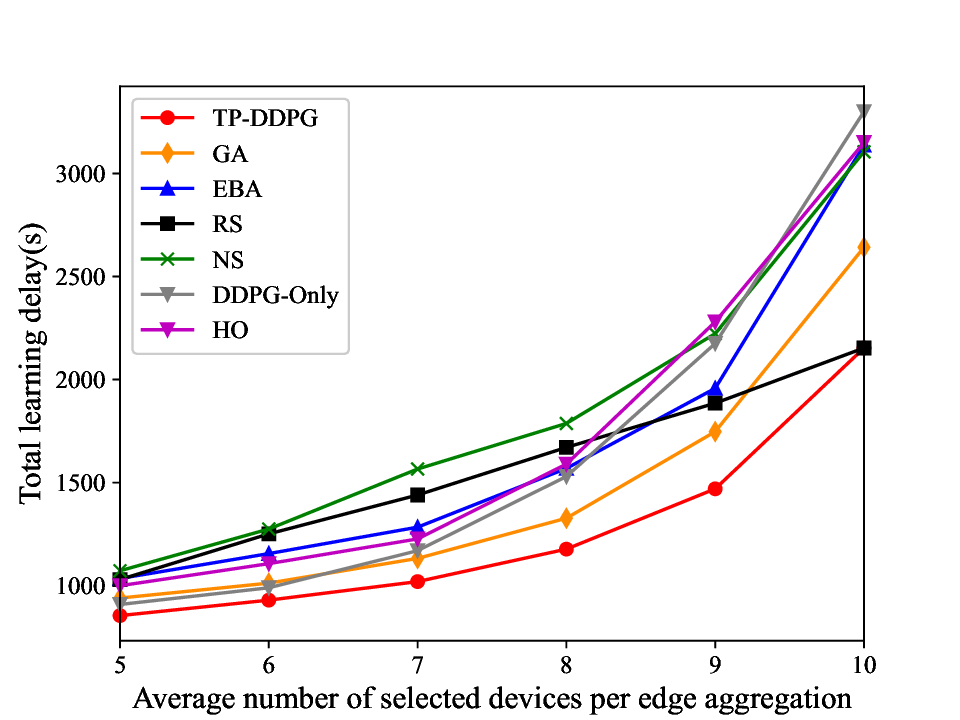}
    \caption{Learning delay versus the average number of scheduled clients per edge aggregation round.}
    \label{fig:delay_num}
\end{figure}

Fig.~\ref{fig:delay_num} plots the total learning delay versus the average number of selected clients per edge aggregation round under the different schemes. It is shown that as long as the number of selected clients remains the same, the proposed TP-DDPG algorithm always completes its FL task with the shortest latency.
The improvement of the TP-DDPG increases as the number of selected clients grows, compared with the NS, BEA, GA, and DDPG-Only algorithms. For instance, as the selected clients increase from five to ten, the additional delay required by the DDPG-Only increases from 6.3\% to 53.3\%, compared to the TP-DDPG algorithm. When all clients participate in each edge aggregation (i.e., the number of selected clients is 10), the RS is equivalent to the proposed TP-DDPG algorithm and produces the same delay.

\begin{figure}
    \centering
    \includegraphics[scale=0.5]{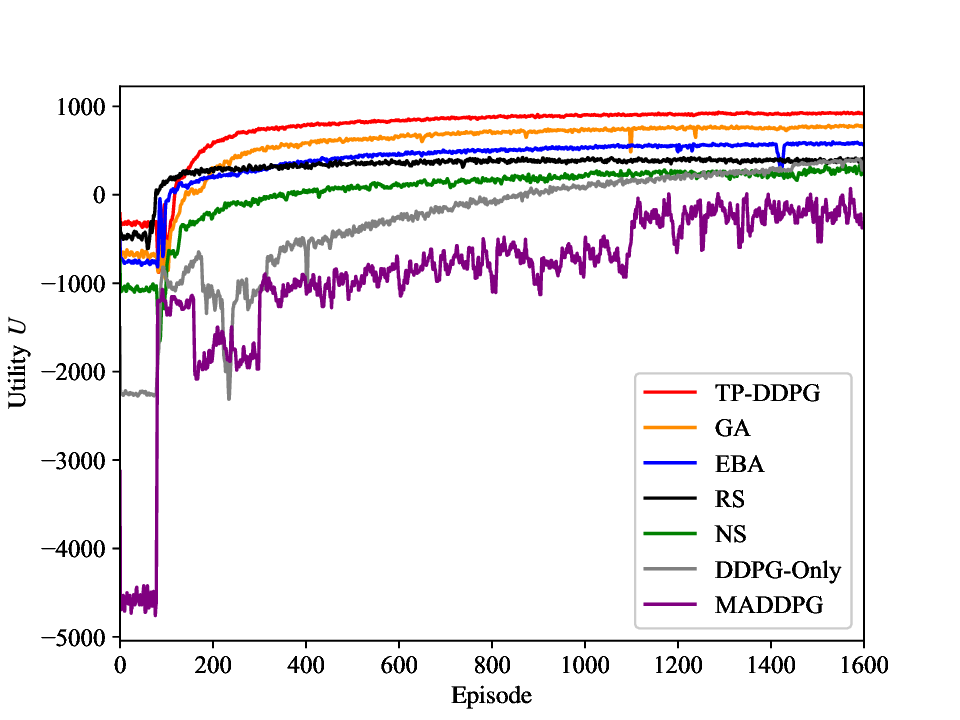}
    \caption{System utility $U$ as the number of episode grows when $\lambda =0.35$.}
    \label{fig:reward_epoch}
\end{figure}

Fig.~\ref{fig:reward_epoch} plots the system utility $U$ as the number of episodes grows, to show the convergence of all algorithms. We see that the utilities of all algorithms are relatively small and stable in the first 80 episodes. This is because the parameters of the DDPG models are randomly initialized and would not be updated until the DDPG memory is full. The proposed TP-DDPG algorithm converges to the largest utility within 2500 episodes. In other words, the algorithm can schedule more clients to join the FL task by adaptive client selection and association. In other words, better learning accuracy can be achieved within a shorter delay. We also see that the convergence rate of the DDPG-Only is slow due to the large dimension of its state and action spaces. 

As shown in Fig.~\ref{fig:reward_epoch}, although the state and action spaces are smaller for each DDPG agent, the MADDPG algorithm converges slower to a lower system utility than the proposed TP-DDPG. Furthermore, placing multiple DDPG agents at the clients can result in additional energy consumption of the clients. Despite its centralized training mechanism, the TP-DDPG substantially reduces the state and action spaces and achieves better convergence by introducing the new straggler-aware client association and bandwidth allocation algorithm, i.e., SCABA.

\begin{figure}[t]
    \centering
    \includegraphics[scale=0.5]{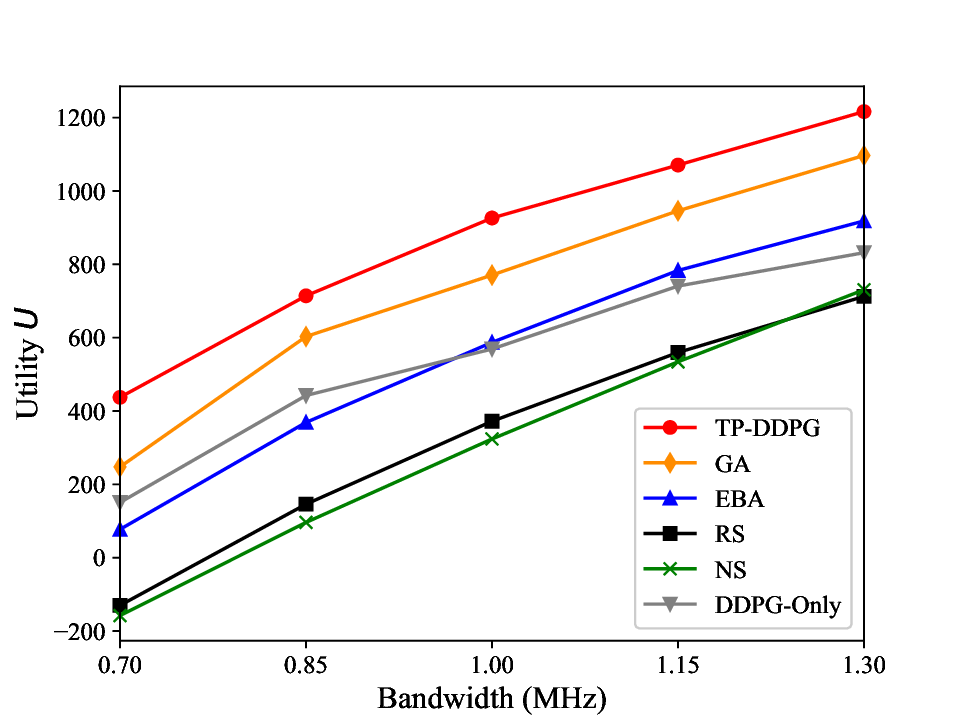}
    \caption{System utility $U$ versus available Bandwidth.}
    \label{fig:obj_bandwidth}
\end{figure}

\begin{figure}[t]
    \centering
    \includegraphics[scale=0.5]{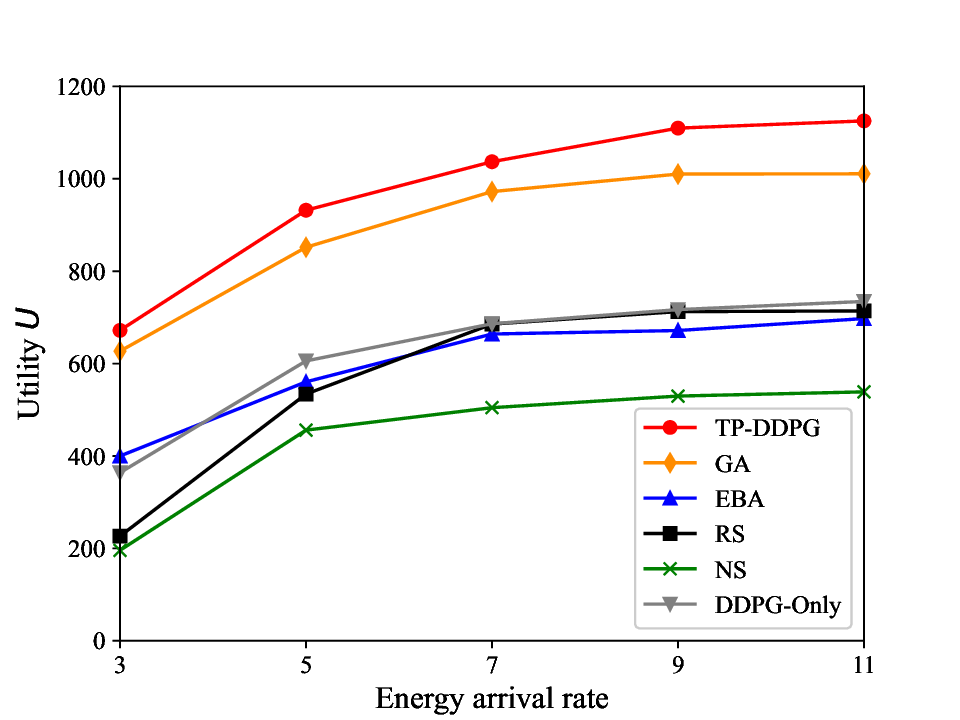}
    \caption{System utility $U$ versus energy arrival rate.}
    \label{fig:obj_energy}
\end{figure}

Fig. \ref{fig:obj_bandwidth} depicts how the available bandwidth affects the performance of the system. As the bandwidth of the system increases, the system utilities of all algorithms grow. This is because a larger bandwidth leads to a higher transmission rate of local model uploading, thus resulting in a shorter transmission delay. It is noted that the utility gain caused by the increased bandwidth of the DDPG-Only is smaller than that of the TP-DDPG. As the available bandwidth grows, the DDPG-Only faces a larger action space for the bandwidth allocation, thus deteriorating the learning performance.

Fig. \ref{fig:obj_energy} shows the system utilities of different algorithms, as the energy arrival rate increases. All schemes show an improvement in system utility when the energy arrival rate increases. The TP-DDPG algorithm always outperforms the benchmarks, as the average energy arrival rate grows from 0.3 W to 1.1 W. With increasing energy arrival rates, the utility $U$ increases slowly and eventually stabilizes. This is because  excessively harvested energy would cause the battery to overflow under the limited battery capacity.

\begin{figure}[t]
\centering
\includegraphics[scale=0.5]{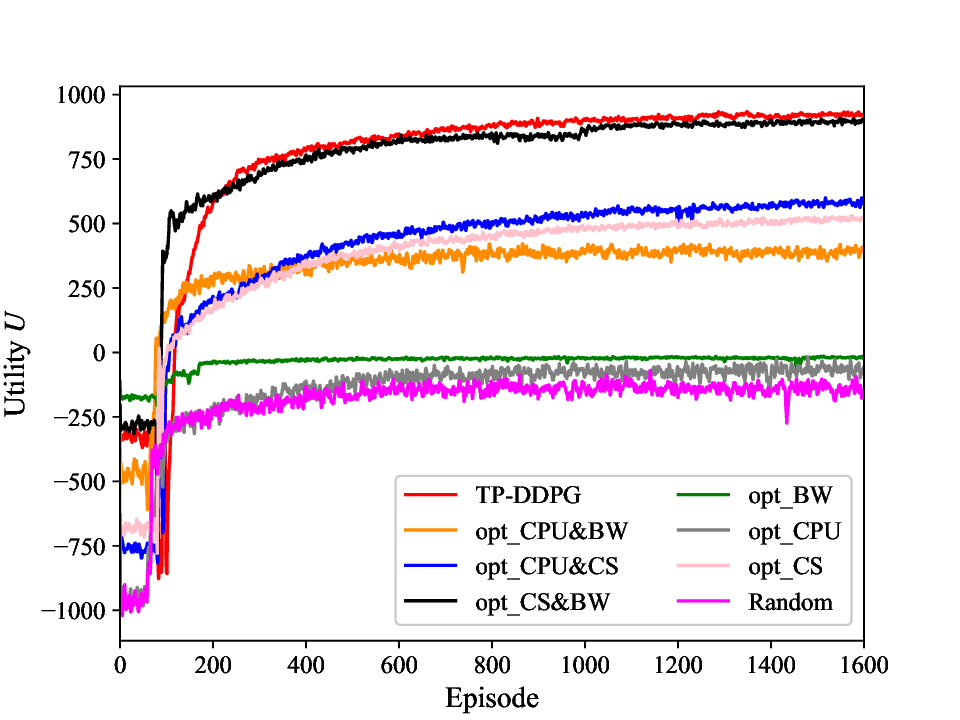}
\caption{An ablation study of the system utility $U$ as the number of episodes grows.}
\label{fig:ablation}
\end{figure}

An ablation study is conducted to optimize client selection (CS), bandwidth (BW), and CPU frequency separately (as compared to the proposed joint optimization in TP-DDPG), as shown in Fig.~\ref{fig:ablation}. It is evident that when only one of the three decisions is optimized while setting the other two randomly, optimizing client selection outperforms optimizing either bandwidth or CPU frequency alone. When two decisions are taken into account, jointly optimizing client selection and bandwidth performs best in improving the utility. 
While the TP-DDPG outperforms all benchmarks by conducting comprehensive optimization across all three decisions, client selection contributes predominantly to the superiority of the method, followed by bandwidth allocation.

% Table \ref{TableII} investigate the changes of energy exhausted times during the training process. We can observe that average battery energy exhausted times per episode of ADS scheme is larger than other schemes in the first 80 episodes due to all clients are selected. As for clients scheduling  schemes, there is a peak after 80 episodes. This is because DRL agent has just began to learn, it only learns that larger transmission power and CPU frequency can get a lower $O_t$ and ignore the energy constraint due to limited interactions with environment. As the episode increases, DRL agent begins to learn the constraint that battery energy cannot be depleted. Finally, when the DDPG network converges, as we expected, the number of battery drains for all schemes is zero.

% To focus on the impact of weight $\lambda$, average number of scheduled clients and delay versus $\lambda$ are shown in Fig. \ref{fig:num_weight}. As can be observed, both of them increase as $\lambda$ increases which means that we can achieve the trade-off between maximizing the number of scheduled clients and minimizing the learning delay by adjusting $\lambda$. 

\begin{figure}[t]
\centering
\subfigure[Test accuracy versus learning delay using MNIST.]{
\includegraphics[width=0.93\linewidth]{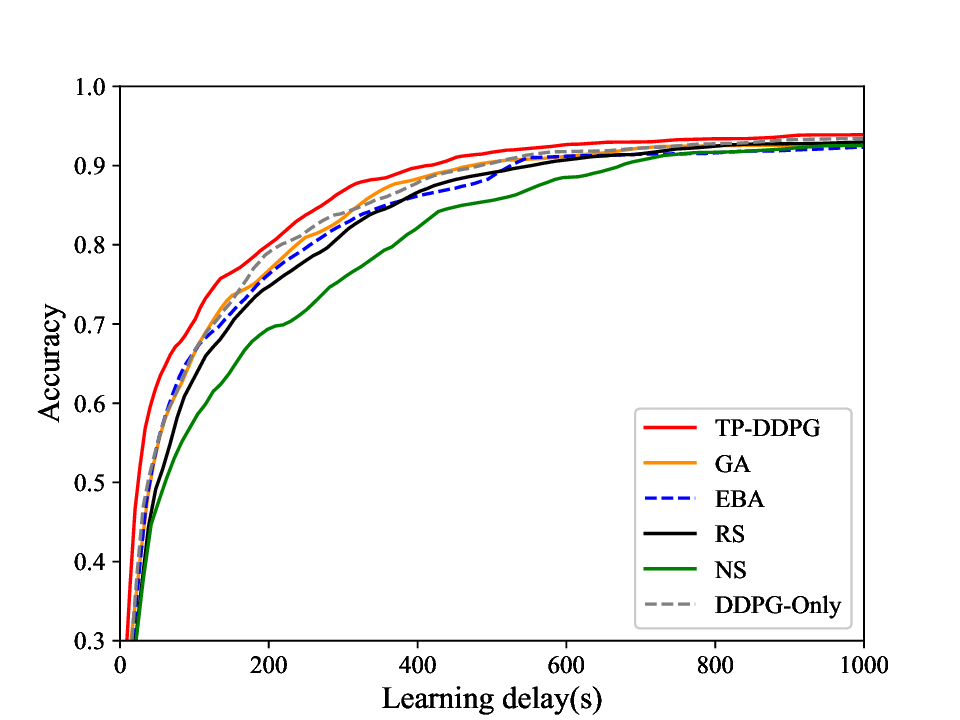}
\label{fig:accuracy_delay}}
\subfigure[Test accuracy versus learning delay using CIFAR-10.]{
\includegraphics[width=0.93\linewidth]{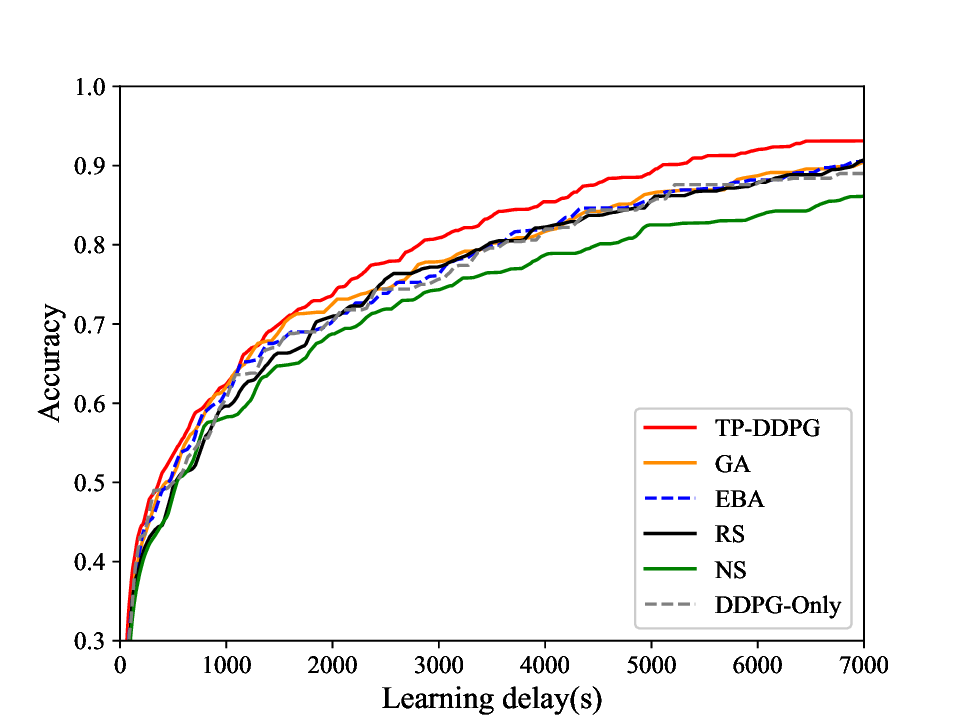}
\label{fig:accuracy_delay_CIF}}
\caption{Test accuracy versus learning delay using different datasets.}
\end{figure}

The test accuracy of the FL models derived under the different algorithms is plotted with respect to the learning delay in Fig. \ref{fig:accuracy_delay}. The test accuracy is the average of ten independent tests. We see that the proposed TP-DDPG always achieves the highest test accuracy. Given a predefined accuracy, the TP-DDPG significantly shortens the learning time. For instance, the TP-DDPG can save up to 39.4\% of the learning time compared to the NS, when the required test accuracy is 0.9. This is because the TP-DDPG changes the resource allocation and client scheduling policy for different edge aggregations, creating a better opportunity to learn new samples from heterogeneous clients than the static NS. In addition, Fig.~\ref{fig:accuracy_delay_CIF} shows the test accuracy of the FL models under
the considered algorithms based on the CIFAR-10 dataset, where each client is
assigned 3,000 heterogeneous training samples with five labels. We see that the proposed TP-DDPG algorithm converges to the highest accuracy of 0.93, while the accuracy is lower than 0.9 for all the benchmarks.

\begin{figure}
    \centering
    \includegraphics[scale=0.5]{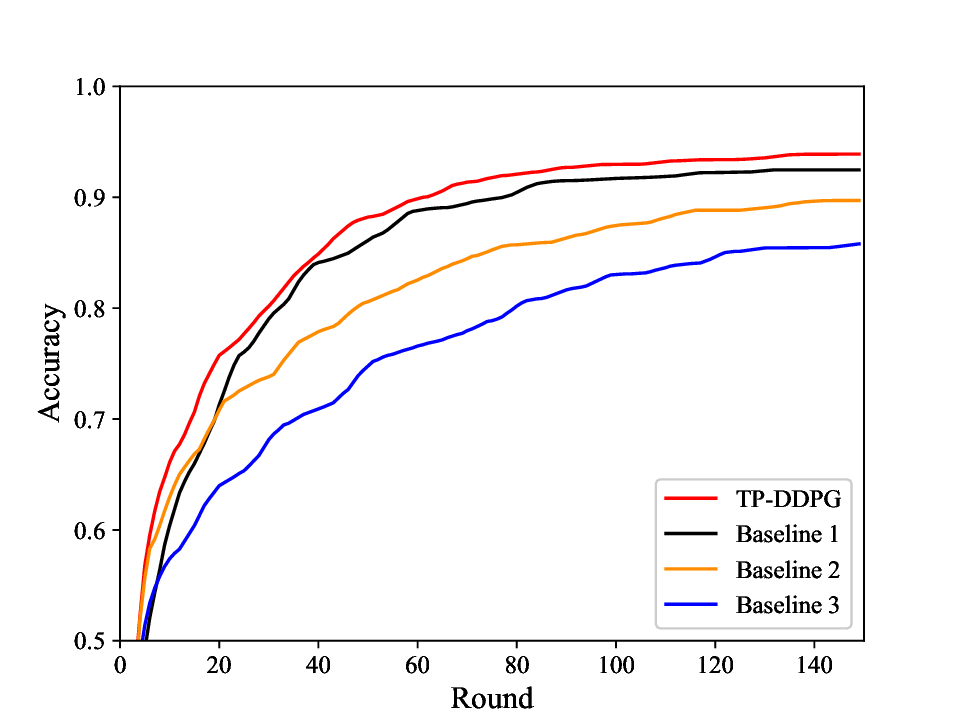}
    \caption{Test accuracy versus cloud aggregation rounds.}
    \label{fig:gap}
\end{figure}

\begin{figure}
    \centering
    \includegraphics[scale=0.5]{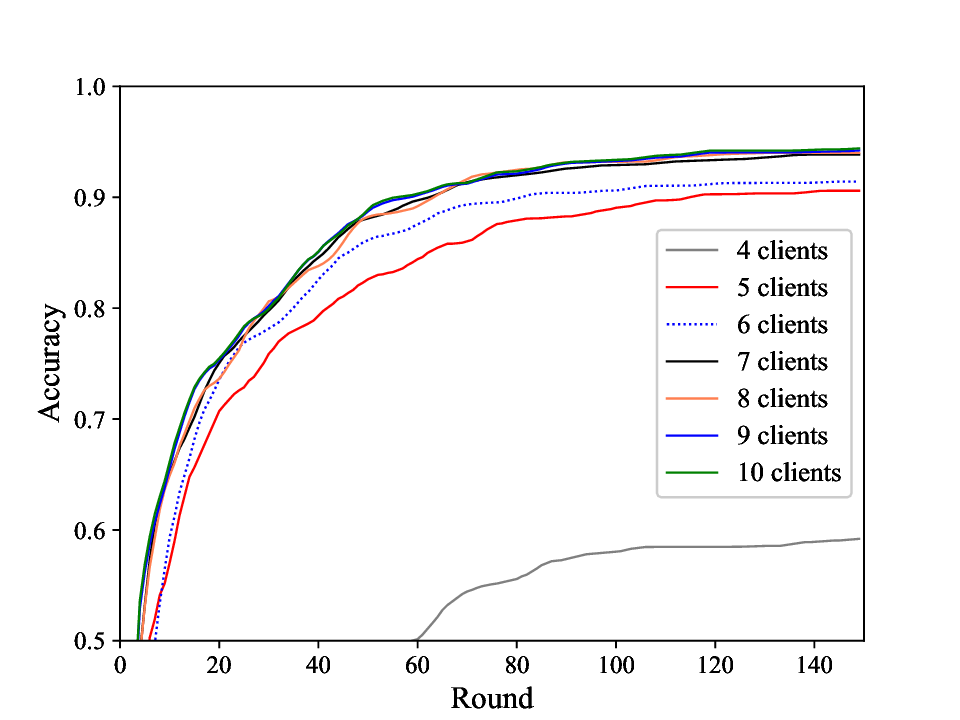}
    \caption{Test accuracy under the TP-DDPG with different numbers of clients.}
    \label{fig:usernumber}
\end{figure}

Fig. \ref{fig:gap} reveals the effects of constraint \eqref{Tf} that defines the maximum round for not selecting a user and the importance-oriented weighting scheme for edge model aggregation in \eqref{imp-weight}. 
%applied for heterogeneous clients on FL performance is shown. 
The proposed TP-DDPG is compared with three baselines; namely, i) Baseline 1, where constraint \eqref{Tf} is removed from problem \eqref{P1};
ii) Baseline 2, where the edge model aggregation follows sample number-based weighting, as done in \cite{luo2020hfel}; and 
iii) Baseline 3, where constraint \eqref{Tf} is removed from problem \eqref{P1} and the edge model aggregation follows the sample number-based weighting. 
Fig. \ref{fig:gap} also shows that the baseline algorithms achieve worse accuracy than the TP-DDPG within the same number of cloud aggregation rounds.  
Without constraining the maximum number of rounds for not selecting a user, Baseline 1 may keep selecting clients with good channel conditions. A client with a critical dataset but poor channel conditions may not be chosen. This renders poor FL accuracy. By applying the importance-oriented weighting in the edge model aggregation, the proposed TP-DDPG effectively deals with the sample heterogeneity of clients and enhances the model accuracy, rather than concentrating only on the sample number of clients, as done in Baseline 2. 

Fig.~\ref{fig:usernumber} plots the test accuracy of the FL model within 150 rounds of cloud aggregations under the proposed TP-DDPG, where the number of clients increases from 4 to 10. It is observed that with the same number of cloud aggregation rounds, the FL model can achieve finer accuracy with more clients participating in learning. A significant accuracy improvement is observed when the number of clients is increased from 4 to 5. Nevertheless, the improvement decreases as the number of clients further increases.
%However, this comes at the expense of edge aggregation latency. For example, compared with Proposal without $F$, the time required for Proposal to complete one edge aggregation has increased from 1.233 s to 1.36 s.

\section{Conclusion}
The learning delay and model accuracy of an FL task in an HFL system with energy harvesting clients were balanced using a joint resource allocation and client scheduling problem. We developed a new TP-DDPG algorithm that optimizes online energy management, computation and communication resource allocation, and client scheduling, adapting to varying wireless channels and renewable energy sources. The algorithm learns the selection of participating clients, and the CPU configurations and transmission powers of the clients. The rest of the decisions, i.e., client association and bandwidth allocation, and the reward of the DDPG are efficiently optimized by the new SCABA algorithm.
%, which evaluates the reward of the DDPG and substantially improves its convergence rate and stability.
%Moreover, to mitigate the learning performance degradation in the case of non-iid, we introduce a parameter $F$ to make sure each device is selected at least once every $F$ edge aggregations. Finally, 
Experimental results showed that the proposed TP-DDPG algorithm can achieve a higher test accuracy with a lower learning latency than the existing benchmarks.

\label{sec:con}

\bibliographystyle{IEEEtran}
\bibliography{references}

\end{document}